\documentclass[10pt,twocolumn,letterpaper]{article}

\usepackage[pagenumbers]{cvpr} 

\usepackage{graphicx}
\usepackage{amsmath}
\usepackage{amssymb}
\usepackage{booktabs}
\usepackage{multirow}

\usepackage{booktabs}
\usepackage{multicol}
\usepackage{color}
\usepackage{caption}
\usepackage{hhline}
\usepackage{pifont}
\usepackage{threeparttable}
\usepackage{makecell}
\usepackage{algorithm} 
\usepackage{listings}
\usepackage{amsfonts,amssymb}
\usepackage{amsmath,amsthm,bm}
\usepackage{graphicx}
\usepackage{tabularx}
\usepackage{lipsum}
\usepackage[accsupp]{axessibility}  

\newcommand\blfootnote[1]{%
  \begingroup
  \renewcommand\thefootnote{}\footnote{#1}%
  \addtocounter{footnote}{-1}%
  \endgroup
}

\usepackage[pagebackref,breaklinks,colorlinks]{hyperref}

\usepackage[capitalize]{cleveref}
\crefname{section}{Sec.}{Secs.}
\Crefname{section}{Section}{Sections}
\Crefname{table}{Table}{Tables}
\crefname{table}{Tab.}{Tabs.}

\def\confName{CVPR}
\def\confYear{2023}

\begin{document}

\title{MobileOne: An Improved One millisecond Mobile Backbone}

\author{
Pavan Kumar Anasosalu Vasu$^\dagger$ \quad James Gabriel \quad Jeff Zhu \quad Oncel Tuzel \quad Anurag Ranjan$^\dagger$ \\ \\ Apple
}
\maketitle

\begin{abstract}
Efficient neural network backbones for mobile devices are often optimized for metrics such as FLOPs or parameter count. However, these metrics may not correlate well with latency of the network when deployed on a mobile device. 
Therefore, we perform extensive analysis of different metrics by deploying several mobile-friendly networks on a mobile device. 
We identify and analyze architectural and optimization bottlenecks in recent efficient neural networks and provide ways to mitigate these bottlenecks. To this end, we design an efficient backbone \textit{MobileOne}, with variants achieving an inference time under 1 ms on an iPhone12 with 75.9\% top-1 accuracy on ImageNet. We show that MobileOne achieves state-of-the-art performance within the efficient architectures while being many times faster on mobile. Our best model obtains similar performance on ImageNet as MobileFormer while being 38$\times$ faster. 
Our model obtains 2.3\% better top-1 accuracy on ImageNet than EfficientNet at similar latency. Furthermore, we show that our model generalizes to multiple tasks -- image classification, object detection, and semantic segmentation with significant improvements in latency and accuracy as compared to existing efficient architectures when deployed on a mobile device. Code and models are available at \url{https://github.com/apple/ml-mobileone}
\blfootnote{corresponding authors: {\{panasosaluvasu, anuragr\}@apple.com}}

\end{abstract}

\section{Introduction}
\begin{figure*}
  \begin{subfigure}[c]{.58\linewidth}
    \centering
    \includegraphics[width=\linewidth]{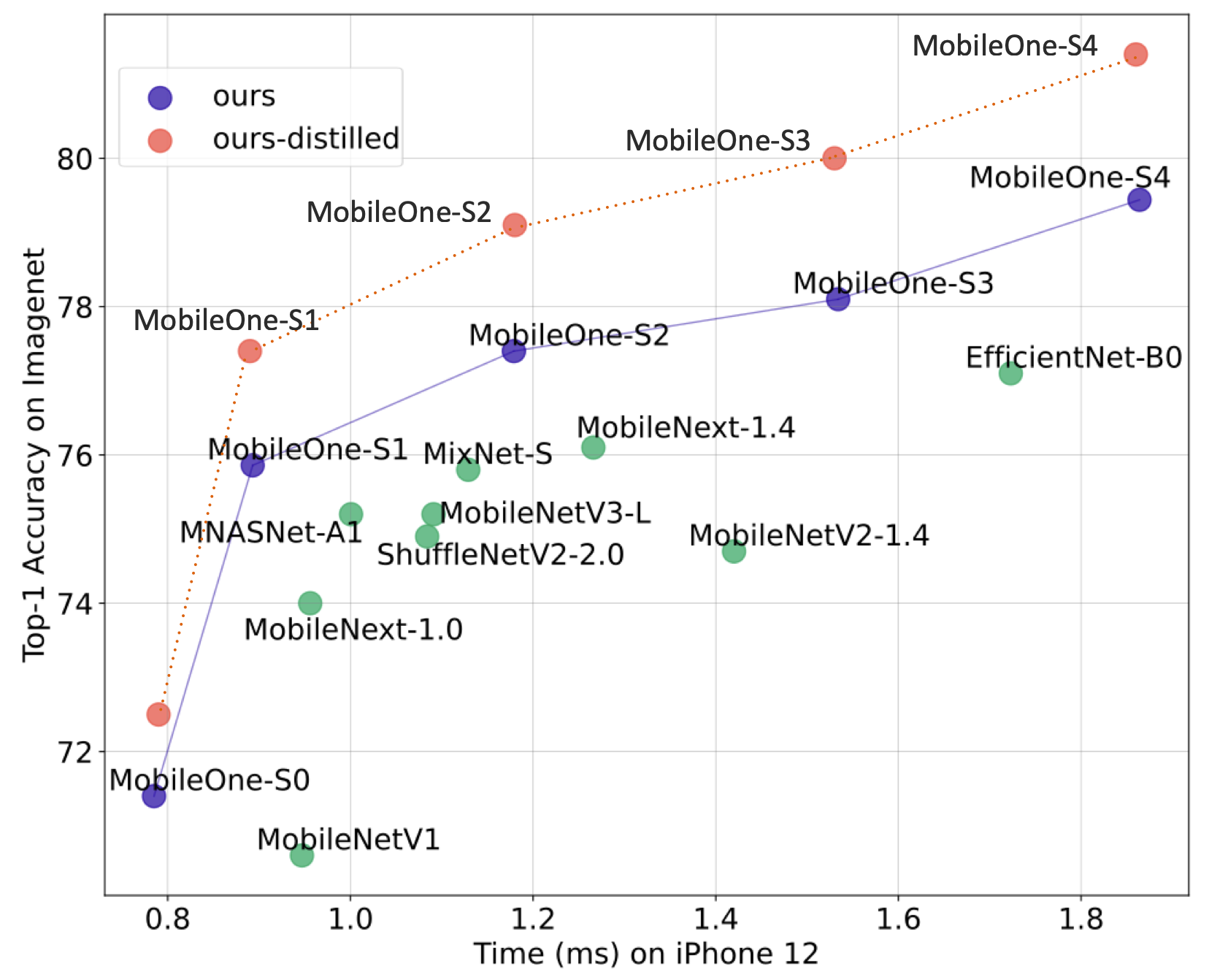}%
    \caption
      {%
            Top 1 accuracy vs Latency on iPhone 12.%
            \label{fig:top1_vs_latency_zoomed}
      }%
  \end{subfigure}\hfill
  \begin{tabular}[c]{@{}c@{}}
    \begin{subfigure}[c]{.40\linewidth}
      \centering
      \includegraphics[width=\linewidth]{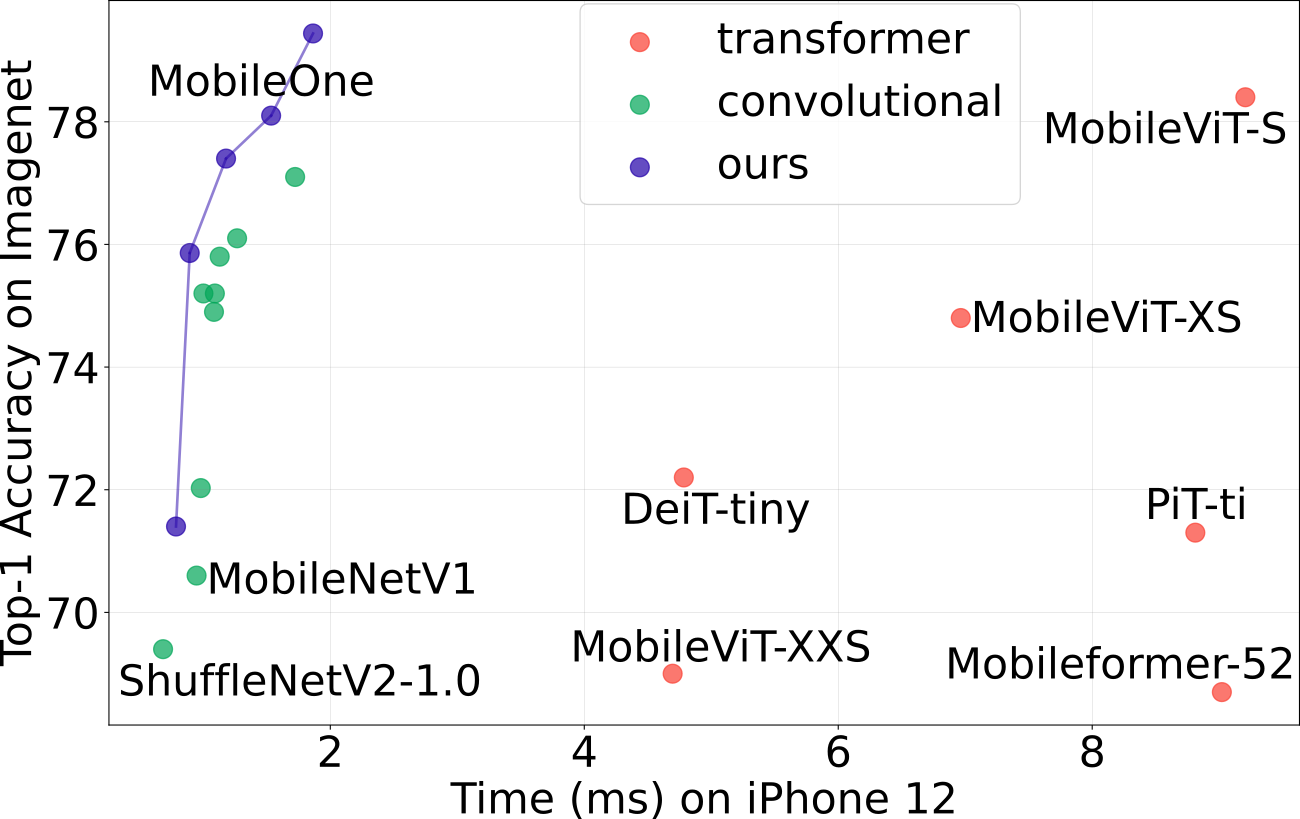}%
      \caption
        {%
          Zoomed out (a)%
            \label{fig:top1_vs_latency}
        }%
    \end{subfigure}\\
    \noalign{\bigskip}%
    \begin{subfigure}[c]{.40\linewidth}
      \centering
      \includegraphics[width=\linewidth,page=2]{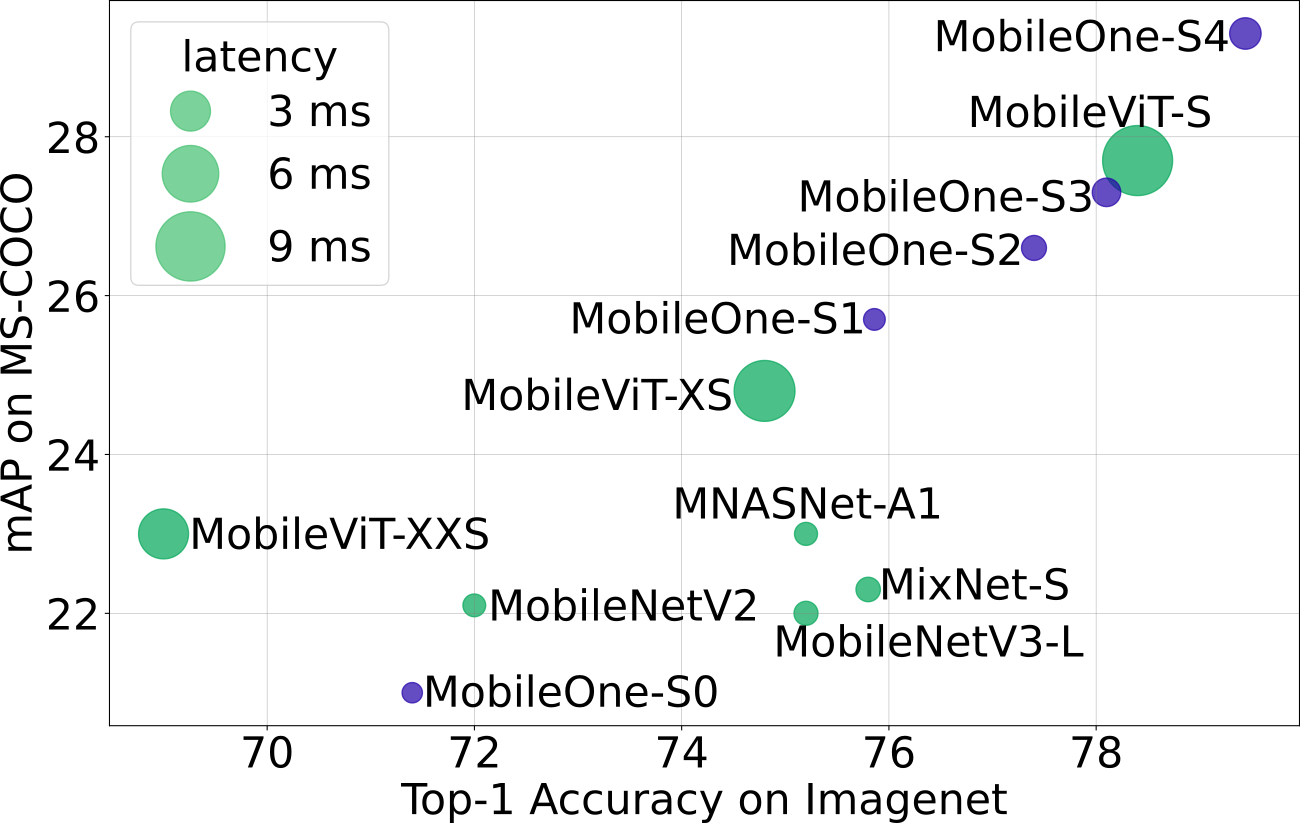}%
      \caption
        {%
          Top-1 accuracy vs mAP.
            \label{fig:top1_vs_map}
        }%
    \end{subfigure}
  \end{tabular}
  \caption
    {%
      We show comparisons of Top-1 accuracy on image classification vs latency on an iPhone 12 (a), and zoomed out area (b) to include recent transformer architectures. We show mAP on object detection vs Top-1 accuracy on image classification in (c) with size of the marker indicating latency of the backbone on iPhone 12. Our models have significantly smaller latency compared to related works. Please refer to supp. mat. for higher resolution figures.
      \label{fig:teaser}%
    }
\end{figure*}

Design and deployment of efficient deep learning architectures for mobile devices has seen a lot of progress~\cite{Howard2017MobileNets, sandler2018mobilenetv2, Howard2019MobilenetV3, Ma_2018_shufflenetv2, mobileformer_cvpr2022, mobilevit_iclr2022}
with consistently decreasing floating-point operations (FLOPs) and parameter count while improving accuracy.
However, these metrics may not correlate well with the efficiency~\cite{dehghani2021efficiency} of the models in terms of latency.
Efficiency metric like FLOPs do not account for memory access cost and degree of parallelism, which can have a non-trivial effect on latency during inference~\cite{Ma_2018_shufflenetv2}.
Parameter count is also not well correlated with latency. For example, sharing parameters leads to higher FLOPS but smaller model size.
Furthermore, parameter-less operations like skip-connections~\cite{he_resnet} or branching~\cite{huang2017densenet,SunXLW19_hrnet} can incur significant memory access costs. This disconnect can get exacerbated when custom accelerators are available in the regime of efficient architectures.

Our goal is to improve the latency cost of efficient architectures while improving their accuracy by identifying key architectural and optimization bottlenecks that affect on-device latency. To identify architectural bottlenecks, we deploy neural networks on an iPhone12 by using CoreML~\cite{coremltools} and benchmark their latency costs. 
To alleviate optimization bottlenecks, we decouple train-time and inference-time architectures, i.e. using a linearly over-parameterized model at train-time and re-parameterizing the linear structures at inference~\cite{Ding_2019_ICCV,Ding_2021_repvgg,ding2021diverse}. We further alleviate optimization bottleneck by dynamically relaxing regularization throughout training to prevent the already small models from being over-regularized. 

Based on our findings on the key bottlenecks, we design a novel architecture \emph{MobileOne}, variants of which run under 1 ms on an iPhone12 achieving state-of-the-art accuracy within efficient architecture family while being significantly faster on the device. Like prior works on structural re-parameterization~\cite{Ding_2019_ICCV,Ding_2021_repvgg,ding2021diverse}, MobileOne introduces linear branches at train-time which get re-parameterized at inference. However, a key difference between our model and prior structural re-parameterization works is the introduction of trivial over-parameterization branches, which provides further improvements in low parameter regime and model scaling strategy. At inference, our model has simple feed-forward structure without any branches or skip-connections. Since this structure incurs lower memory access cost, we can incorporate wider layers in our network which boosts representation capacity as demonstrated empirically in Table~\ref{tab:compare_imagenet}.
For example, MobileOne-S1 has 4.8M parameters and incurs a latency of 0.89ms, while MobileNet-V2~\cite{sandler2018mobilenetv2} has 3.4M (29.2\% less than MobileOne-S1) parameters and incurs a latency of 0.98ms. At this operating point, MobileOne attains 3.9\% better top-1 accuracy than MobileNet-V2.  

MobileOne achieves significant improvements in latency compared to efficient models in literature while maintaining the accuracy on several tasks -- image classification, object detection, and semantic segmentation. As shown in Figure ~\ref{fig:top1_vs_latency}, MobileOne performs better than MobileViT-S~\cite{mobilevit_iclr2022} while being 5 $\times$ faster on image classification. As compared to EfficientNet-B0~\cite{efficientnet_v1_plmr}, we achieve 2.3\% better top-1 accuracy on ImageNet~\cite{deng2009imagenet} with similar latency costs (see Figure~\ref{fig:top1_vs_latency_zoomed}). Furthermore, as seen in Figure~\ref{fig:top1_vs_map}, MobileOne models not only perform well on ImageNet, they also generalize to other tasks like object detection. Models like MobileNetV3-L~\cite{Howard2019MobilenetV3} and MixNet-S~\cite{mixconv_bmvc_2019} improve over MobileNetV2 on ImageNet, but those improvements do not translate to object detection task. As shown in Figure~\ref{fig:top1_vs_map}, MobileOne 
shows better generalization across tasks. For object detection on MS-COCO~\cite{mscoco_lin_2014}, best variant of MobileOne outperforms best variant MobileViT by 6.1\% and MNASNet by 27.8\%.
For semantic segmentation, on PascalVOC~\cite{pascal-voc-2012} dataset, best variant of MobileOne outperforms best variant MobileViT by 1.3\% and on ADE20K~\cite{ade20k} dataset, best variant of MobileOne outperforms MobileNetV2 by 12.0\%. In summary, our contributions are as follows:
\begin{itemize}
    \item We introduce \emph{MobileOne}, a novel architecture that runs within 1 ms on a mobile device and achieves state-of-the-art accuracy on image classification within efficient model architectures. The performance of our model also generalizes to a desktop CPU and GPU.
    \item We analyze performance bottlenecks in activations and branching that incur high latency costs on mobile in recent efficient networks.
    \item We analyze the effects of train-time re-parameterizable branches and dynamic relaxation of regularization in training. In combination, they help alleviating optimization bottlenecks encountered when training small models.
    \item We show that our model generalizes well to other tasks -- object detection and semantic segmentation while outperforming recent state-of-the-art efficient models.
\end{itemize}
We will release our trained networks and code for research purposes. We will also release the code for iOS application to enable benchmarking of networks on iPhone.

\section{Related Work}
Designing a real-time efficient neural network involves a trade-off between accuracy and performance. Earlier methods like SqueezeNet~\cite{squeezenet_2016} and more recently MobileViT~\cite{mobilevit_iclr2022}, optimize for parameter count and a vast majority of methods like MobileNets~\cite{Howard2017MobileNets,sandler2018mobilenetv2}, MobileNeXt~\cite{mobilenext_eccv2020}, ShuffleNet-V1~\cite{zhang2018shufflenet},  GhostNet~\cite{Han_ghostnet_2020_CVPR}, MixNet~\cite{mixconv_bmvc_2019} focus on optimizing for the number of floating-point operations (FLOPs). 
EfficientNet~\cite{efficientnet_v1_plmr} and TinyNet~\cite{tinynets_neurips} study the compound scaling of depth, width and resolution while optimizing FLOPs. 
Few methods like MNASNet~\cite{mnasnet_cvpr}, MobileNetV3~\cite{Howard2019MobilenetV3} and ShuffleNet-V2~\cite{Ma_2018_shufflenetv2} optimize directly for latency. Dehghani et al.~\cite{dehghani2021efficiency} show that FLOPs and parameter count are not well correlated with latency. Therefore, our work focuses on improving on-device latency while improving the accuracy. 

Recently, ViT~\cite{dosovitskiy2020image} and ViT-like architectures~\cite{Touvron_2021_ICCV} have shown state-of-the-art performance on ImageNet dataset. Different designs like ViT-C~\cite{vitc_arxiv}, CvT~\cite{wu2021cvt}, BoTNet~\cite{botnet_cvpr2021}, ConViT~\cite{d2021convit} and PiT~\cite{heo2021pit} have been explored to incorporate biases using convolutions in ViT. More recently, MobileFormer~\cite{mobileformer_cvpr2022} and MobileViT~\cite{mobilevit_iclr2022} were introduced to get ViT-like performance on a mobile platform. MobileViT optimizes for parameter count and MobileFormer optimizes for FLOPs and outperforms efficient CNNs in low FLOP regime. However, as we show in subsequent sections that low FLOPs does not necessarily result in low latency. We study key design choices made by these methods and their impact on latency. 

Recent methods also introduce new architecture designs and custom layers to improve accuracy for mobile backbones. MobileNet-V3~\cite{Howard2019MobilenetV3}, introduces an optimized activation function -- Hard-Swish for a specific platform. However, scaling such functions to different platforms may be difficult.
 
Therefore, our design uses basic operators that are already available across different platforms. ExpandNets~\cite{NEURIPS2020_expandnets}, ACNet~\cite{Ding_2019_ICCV} and DBBNet~\cite{ding2021diverse}, propose a drop-in replacement for a regular convolution layer in recent CNN architectures and show improvements in accuracy.
RepVGG~\cite{Ding_2021_repvgg} introduces re-parameterizable skip connections which is beneficial to train VGG-like model to better performance.
These architectures have linear branches at train-time that get re-parameterized to simpler blocks at inference. We build on these re-parametrization works and introduce trivial over-parameterization branches thereby providing further improvements in accuracy.

\section{Method}
\label{sec:method}
In this section, we analyse the correlation of popular metrics -- FLOPs and parameter count  -- with latency on a mobile device. We also evaluate how different design choices in architectures effect the latency on the phone. Based on the evaluation, we describe our architecture and training algorithm.

\subsection{Metric Correlations}
The most commonly used cost indicators for comparing the size of two or more models are parameter count and FLOPs~\cite{dehghani2021efficiency}. However, they may not be well correlated with latency in real-world mobile applications.
Therefore, we study the correlation of latency with FLOPS and parameter count for benchmarking efficient neural networks. We consider recent models and use their Pytorch implementation to convert them into ONNX format~\cite{bai2019onnx}. We convert each of these models to coreml packages using Core ML Tools~\cite{coremltools}. 
We then develop an iOS application to measure the latency of the models on an iPhone12.

\begin{figure}
    \centering
    \includegraphics[width=0.94\linewidth]{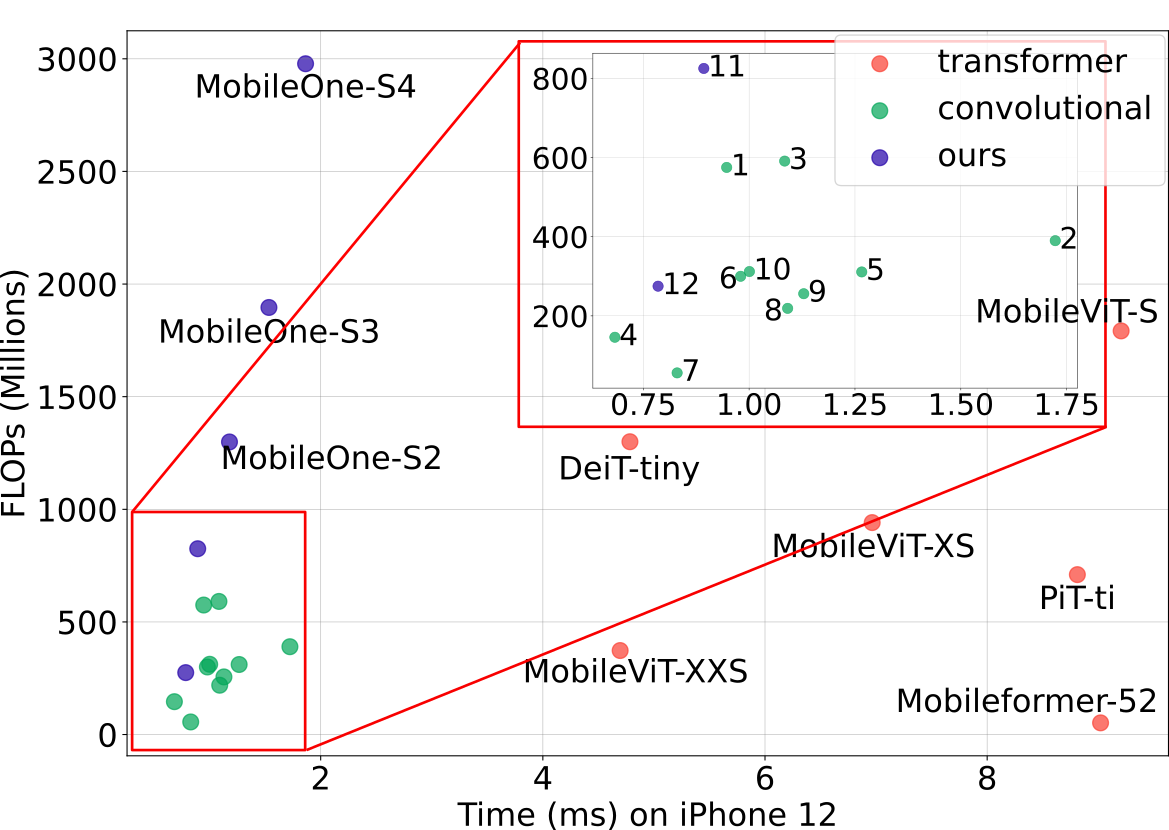} \\ 
    \includegraphics[width=0.94\linewidth]{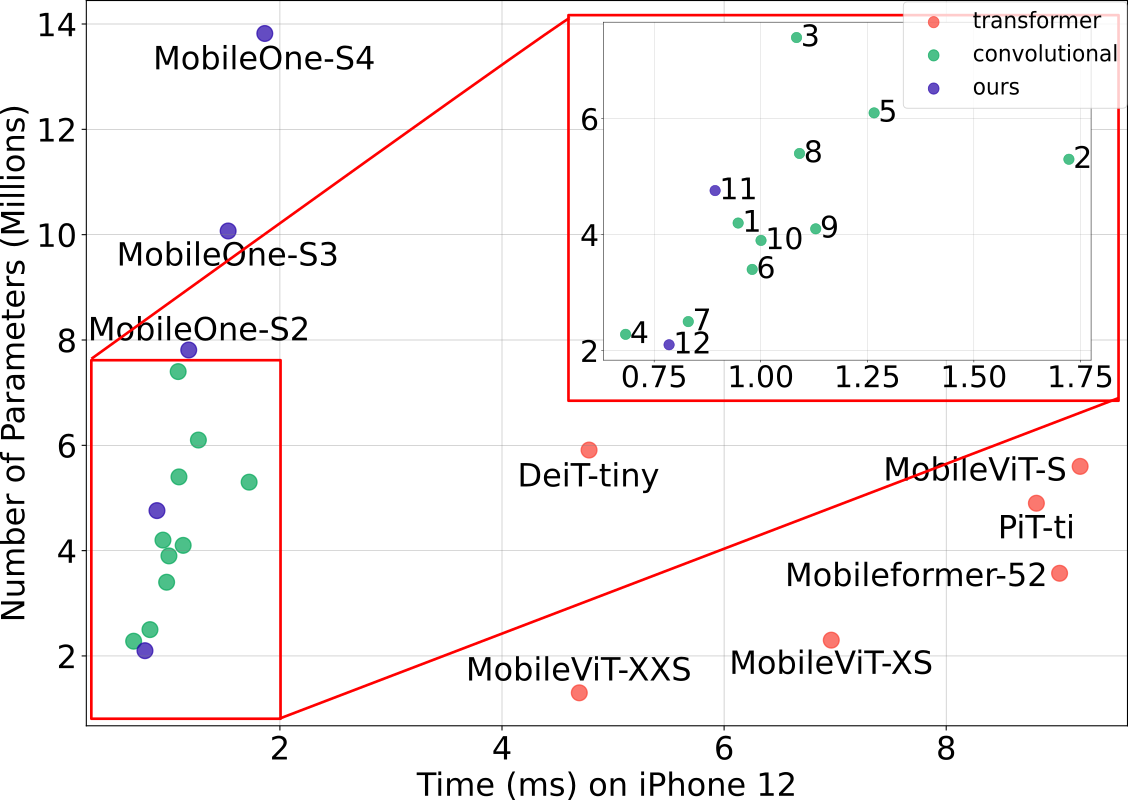} 
    
    \scalebox{0.7}{
    \begin{tabular}{cccc}
    \\
        \hline 
         1 & 2 & 3 & 4 \\
         MobileNetV1 & EfficientNet-B0 & ShuffleNetV2-2.0 &
        ShuffleNetV2-1.0 \\ \hline
        5 & 6 & 7 & 8  \\
        MobileNext-1.4 & MobileNetV2  & MobileNetV3-S & MobileNetV3-L \\
        \hline
        9 & 10 & 11 & 12 \\
        MixNet-S &
       MNASNet-A1 & MobileOne-S1 & MobileOne-S0 \\
       \hline
    \end{tabular}}
    \caption{Top: FLOPs vs Latency on iPhone12. Bottom: Parameter Count vs Latency on iPhone 12. We indicate some networks using numbers as shown in the table above.}
    \label{fig:latency_graph}
\end{figure}

We plot latency vs. FLOPs and latency vs. parameter count as shown in Figure~\ref{fig:latency_graph}.
We observe that many models with higher parameter count can have lower latency. We observe a similar plot between FLOPs and latency. Furthermore, we note the convolutional models such as MobileNets~\cite{sandler2018mobilenetv2, Ma_2018_shufflenetv2, efficientnet_v2_quoc} have lower latency for similar FLOPs and parameter count than their transformer counterparts~\cite{Touvron_2021_ICCV, mobileformer_cvpr2022, mobilevit_iclr2022}.
We also estimate the Spearman rank correlation~\cite{zar2005spearman} in Table~\ref{tab:correlation_coeff}a. 
We find that latency is moderately correlated with FLOPs and weakly correlated with parameter counts for efficient architectures on a mobile device. This correlation is even lower on a desktop CPU.

\begin{table}
    \centering
    \footnotesize 
    \scalebox{0.85}{
    \begin{tabular}{lcccc}
    \toprule
        & \multicolumn{2}{c}{\textbf{FLOPs}} & \multicolumn{2}{c}{\textbf{Parameters}} \\
        \textbf{Type} & \textbf{corr.} & \textbf{p-value} & \textbf{corr.} & \textbf{p-value} \\
    \midrule
         \textbf{Mobile Latency} & 0.47 & 0.03 & 0.30 & 0.18\\
         \textbf{CPU Latency} & 0.06 & 0.80 & 0.07 & 0.77\\
    \bottomrule
    \end{tabular}
    }
    \caption{Spearman rank correlation coeff. between latency-flops.}
    \label{tab:correlation_coeff}
\end{table}

\subsection{Key Bottlenecks}

\paragraph{Activation Functions}
To analyze the effect of activation functions on latency, we construct a 30 layer convolutional neural network and benchmark it on iPhone12 using different activation functions, commonly used in efficient CNN backbones. All models in Table~\ref{tab:activation_latency} have the same architecture except for activations, but their latencies are drastically different. This can be attributed to synchronization costs mostly incurred by recently introduced activation functions like SE-ReLU~\cite{hu2018senet}, Dynamic Shift-Max~\cite{Li_2021_ICCV_micronet} and DynamicReLUs~\cite{chen2020dynamic}.
DynamicReLU and Dynamic Shift-Max have shown significant accuracy improvement in extremely low FLOP models like MicroNet~\cite{Li_2021_ICCV_micronet}, but, the latency cost of using these activations can be significant. 
Therefore we use only ReLU activations in MobileOne.

\begin{table}[]
 \vspace{-0.2cm}
 \centering
 \scalebox{0.9}{
 \small
      \begin{tabular}{lc}
            \toprule
            \textbf{Activation Function} & \textbf{Latency (ms)}  \\
            \midrule
            ReLU~\cite{agarap2018deep_relu} & 1.53  \\
            GELU~\cite{hendrycks2016gelu} & 1.63  \\
            SE-ReLU~\cite{hu2018senet} & 2.10 \\
            SiLU~\cite{silu_2017_elfwing} & 2.54  \\
            Dynamic Shift-Max~\cite{Li_2021_ICCV_micronet} & 57.04  \\
            DynamicReLU-A~\cite{chen2020dynamic} & 273.49  \\
            DynamicReLU-B~\cite{chen2020dynamic} & 242.14  \\
            \bottomrule
        \end{tabular}
        }
        \vspace{-0.2cm}
      \captionof{table}{Comparison of latency on mobile device of different activation functions in a 30-layer convolutional neural network.}\label{tab:activation_latency} 
      \vspace{-0.2cm}
\end{table}

\paragraph{Architectural Blocks}
Two of the key factors that affect runtime performance are memory access cost and degree of parallelism~\cite{Ma_2018_shufflenetv2}. 
Memory access cost increases significantly in multi-branch architectures as activations from each branch have to be stored to compute the next tensor in the graph. 
Such memory bottlenecks can be avoided if the network has smaller number of branches.
Architectural blocks that force synchronization like global pooling operations used in Squeeze-Excite block~\cite{hu2018senet} also affect overall run-time due to synchronization costs.
To demonstrate the hidden costs like memory access cost and synchronization cost, we ablate over using skip connections and squeeze-excite blocks in a 30 layer convolutional neural network. In Table~\ref{table:arch_blocks_vs_latency}b, we show how each of these choices contribute towards latency. Therefore we adopt an architecture with no branches at inference, which results in smaller memory access cost. In addition, we limit the use of Squeeze-Excite blocks to our biggest variant in order to improve accuracy. 

\begin{table}
\centering
\scalebox{0.8}{
    \begin{tabular}{lccc}
        \toprule
        \textbf{Architectural} & \multirow{2}{*}{Baseline}  & {+ Squeeze} & {+ Skip}  \\
        \textbf{Blocks }& & {Excite~\cite{hu2018senet}
        } & { Connections~\cite{He2015}} \\
        \midrule
        \textbf{Latency (ms)} & 1.53 & 2.10 & 2.62 \\
        \bottomrule
    \end{tabular}}
    \caption{Ablation on latency of different architectural blocks in a 30-layer convolutional neural network.}\label{table:arch_blocks_vs_latency}
\end{table}

\subsection{MobileOne Architecture}\label{section:arch}
Based on the our evaluations of different design choices, we develop the architecture of MobileOne. Like prior works on structural re-parameterization~\cite{NEURIPS2020_expandnets,Ding_2019_ICCV,Ding_2021_repvgg,ding2021diverse}, the train-time and inference time architecture of MobileOne is different. In this section, we introduce the basic block of MobileOne and the model scaling strategy used to build the network.

\begin{figure}
    \centering
    \includegraphics[width=0.75\linewidth]{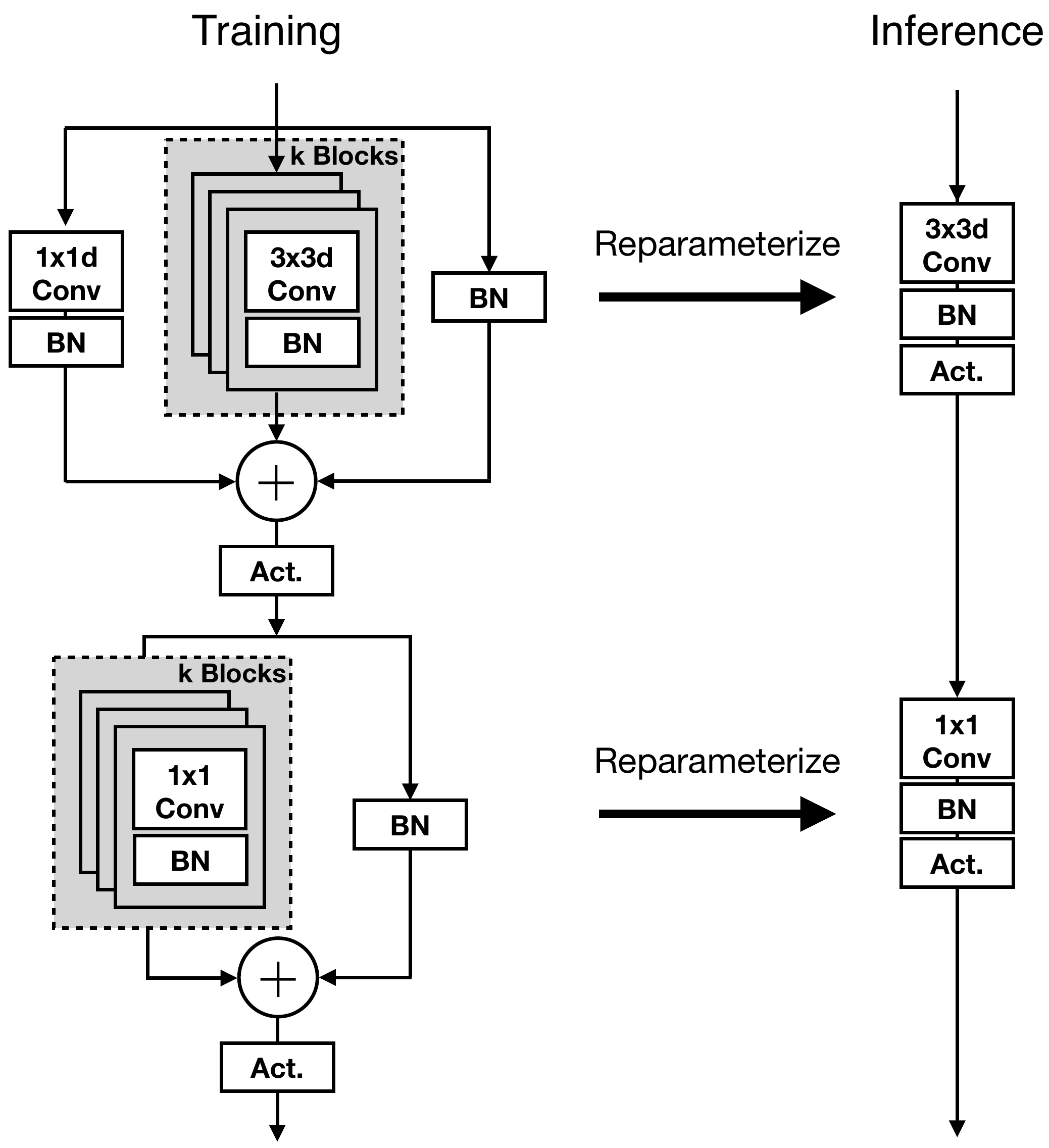}
    \captionof{figure}{MobileOne block has two different structures at train time and test time. Left: Train time MobileOne block with reparameterizable branches. Right: MobileOne block at inference where the branches are reparameterized. Either ReLU or SE-ReLU is used as activation. The trivial over-parameterization factor $k$ is a hyperparameter which is tuned for every variant.}\label{fig:mobileone_blocks}
 \end{figure}

\paragraph{MobileOne Block}
MobileOne blocks are similar to blocks introduced in~\cite{NEURIPS2020_expandnets,Ding_2019_ICCV,Ding_2021_repvgg,ding2021diverse}, except that our blocks are designed for convolutional layers that are factorized into depthwise and pointwise layers. Furthermore, we introduce trivial over-parameterization branches which provide further accuracy gains. Our basic block builds on the MobileNet-V1~\cite{Howard2017MobileNets} block of 3x3 depthwise convolution followed by 1x1 pointwise convolutions. We then introduce re-parameterizable skip connection~\cite{Ding_2021_repvgg} with batchnorm along with branches that replicate the structure as shown in Figure~\ref{fig:mobileone_blocks}. The trivial over-parameterization factor $k$ is a hyperparameter which is varied from 1 to 5. We ablate over the choice for $k$ in Table~\ref{tab:compare_traintime_overparam}. At inference, MobileOne model does not have any branches. They are removed using the re-parameterization process described in ~\cite{Ding_2021_repvgg, ding2021diverse}.

 \begin{table}[]
 \centering
 \small
 \scalebox{0.9}{
        \begin{tabular}{lccc}
            \toprule
            \textbf{Model} & \textbf{\# Params.} & \textbf{Top-1} \\
            \midrule
            ExpandNet-CL MobileNetV1~\cite{NEURIPS2020_expandnets}        & 4.2 & 69.4 & \\
            RepVGG-A0~\cite{Ding_2021_repvgg}     & 8.3  & 72.4 \\
            RepVGG-A1~\cite{Ding_2021_repvgg}     & 12.8 & 74.5 \\
            RepVGG-B0~\cite{Ding_2021_repvgg}     & 14.3 & 75.1 \\
            ACNet MobileNetV1~\cite{Ding_2019_ICCV}   & 4.2  & 72.1 \\
            ACNet ResNet18~\cite{Ding_2019_ICCV}      & 11.7 & 71.1 \\
            DBBNet MobileNetV1~\cite{ding2021diverse}  & 4.2  & 72.9 \\
            DBBNet ResNet18~\cite{ding2021diverse}     & 11.7 & 71.0 \\
            \midrule
            \textbf{MobileOne-S0}   & 2.1 & 71.4 \\
            \textbf{MobileOne-S1}   & 4.8 & 75.9 \\
            \textbf{MobileOne-S2}   & 7.8 & 77.4 \\
            \textbf{MobileOne-S3}   & 10.1 & 78.1 \\
            \textbf{MobileOne-S4}   & 14.8 & 79.4 \\
            \bottomrule
        \end{tabular}
      }
      \vspace{-0.2cm}
      \captionof{table}{Comparison of Top-1 Accuracy on ImageNet against recent train time over-parameterization works. Number of parameters listed above is at inference.}\label{tab:compare_traintime_overparam}
      \vspace{-0.2cm}
 \end{table} 
\begin{table}
\vspace{-0.2cm}
\centering
\footnotesize
\scalebox{0.9}{
        \begin{tabular}{lccc}
        \\ 
            \toprule
            \textbf{Re-param.} & \textbf{MobileOne-S0} & \textbf{MobileOne-S1} & \textbf{MobileOne-S3} \\ 
            \midrule
            \textbf{with}    &   71.4         & 75.9 & 78.1 \\
            \textbf{without} &  69.6 &    74.6        & 77.2 \\
            \bottomrule
        \end{tabular} 
        }
        \vspace{-0.2cm}
      \captionof{table}{Effect re-parametrizable branches on Top-1 ImageNet accuracy.}~\label{table:ablate_rep_branches}
      \vspace{-0.4cm}
\end{table}

\begin{table}
    \centering
    \small
    \scalebox{0.9}{
          \begin{tabular}{cccccc}
            \toprule
            \multirow{2}{*}{\textbf{Model}} & \multicolumn{5}{c}{\textbf{Top-1}} \\
            \cmidrule{2-6}
            & k=1 & k=2 & k=3 & k=4 & k-5 \\
            \midrule
            MobileOne-S0    & 70.9 & 70.7 & 71.3 & 71.4 & 71.1 \\
            MobileOne-S1    & 75.9 & 75.7 & 75.6 & 75.6 & 75.2 \\
            \bottomrule
        \end{tabular}
        }
        \vspace{-0.2cm}
      \captionof{table}{Comparison of Top-1 on ImageNet for various values of trivial over-parameterization factor $k$.}~\label{table:ablate_k_rep_branches}
      \vspace{-0.2cm}
\end{table}

For a convolutional layer of kernel size $K$, input channel dimension $C_{in}$ and output channel dimension $C_{out}$, the weight matrix is denoted as $\mathbf{W}' \in \mathbb{R}^{C_{out} \times C_{in} \times K \times K}$ and bias is denoted as $\mathbf{b}' \in \mathbb{R}^{D}$. A batchnorm layer contains accumulated mean $\bm{\mu}$, accumulated standard deviation $\bm{\sigma}$, scale $\bm{\gamma}$ and bias $\bm{\beta}$. Since convolution and batchnorm at inference are linear operations, they can be folded into a single convolution layer with weights $\widehat{\mathbf{W}} = \mathbf{W}' * \frac{\bm{\gamma}}{\bm{\sigma}}$ and bias $\widehat{\mathbf{b}} = (\mathbf{b}' - \bm{\mu}) * \frac{\bm{\gamma}}{\bm{\sigma}} + \bm{\beta}$. Batchnorm is folded into preceding convolutional layer in all the branches. For skip connection the batchnorm is folded to a convolutional layer with identity 1x1 kernel, which is then padded by $K-1$ zeros as described in~\cite{Ding_2021_repvgg}. After obtaining the batchnorm folded weights in each branch, the weights $\mathbf{W} =  \sum_{i}^{M} \widehat{\mathbf{W}}_{i}$ and bias $\mathbf{b}=  \sum_{i}^{M} \widehat{\mathbf{b}}_{i}$ for convolution layer at inference is obtained, where $M$ is the number of branches.  

\begin{figure}
    \centering
    \includegraphics[width=1.0\linewidth]{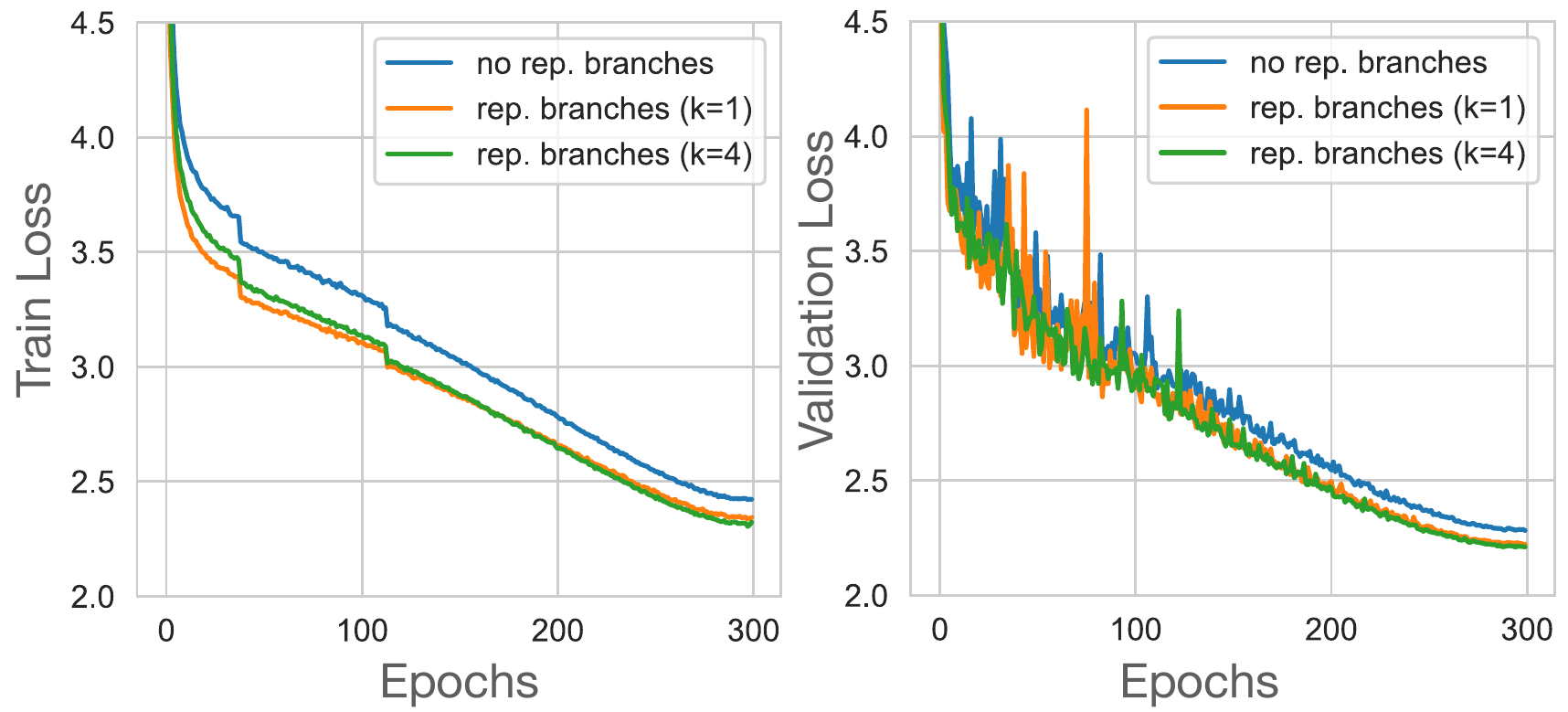}
    \captionof{figure}{Plot of train and validation losses of MobileOne-S0 model. From no branches to adding re-parameterizable branches with $k$=1, leads to 3.4\% lower train loss. Adding more branches ($k$=4) lowers train loss by an additional $\sim$1\%. From no branches to the variant with re-parameterizable branches ($k$=4), validation loss improves by 3.1\%}\label{fig:trainval_loss}
\end{figure}

To better understand the improvements from using train time re-parameterizable branches, we ablate over versions of MobileOne models by removing train-time re-parameterizable branches (see Table~\ref{table:ablate_rep_branches}), while keeping all other training parameters the same as described in Section~\ref{section:experiment_details_imagenet}. Using re-parameterizable branches significantly improves performance. To understand the importance of trivial over-parameterization branches, we ablate over the choice of over-parameterization factor $k$ in Table~\ref{table:ablate_k_rep_branches}. For larger variants of MobileOne, the improvements from trivial over-parameterization starts diminishing. For smaller variant like MobileOne-S0, we see improvements of 0.5\% by using trivial over-parameterization branches. In Figure~\ref{fig:trainval_loss}, we see that adding re-parameterizable branches improves optimization as both train and validation losses are further lowered. 

\begin{table*}
\footnotesize
        \begin{tabular}{ccccccccccc}
          \toprule
          \multirow{2}{*}{\textbf{Stage}} & \multirow{2}{*}{\textbf{Input}} & \multirow{2}{*}{\textbf{\# Blocks}} & \multirow{2}{*}{\textbf{Stride}} & \multirow{2}{*}{\textbf{Block Type}} &
          \multirow{2}{*}{\textbf{\# Channels}} & \multicolumn{5}{c}{\textbf{MobileOne Block Parameters ($\alpha$, $k$, act=ReLU)}} \\
          \cmidrule{7-11}
          & & & & & & \textbf{S0} & \textbf{S1} & \textbf{S2} & \textbf{S3} & \textbf{S4} \\
          \midrule
          1  & $224\times224$ & 1 & 2 & MobileOne-Block & 64$\times\alpha$  & (0.75, 4) & (1.5, 1) & (1.5, 1) & (2.0, 1) & (3.0, 1) \\
          2  & $112\times112$ & 2 & 2 & MobileOne-Block & 64$\times\alpha$  & (0.75, 4) & (1.5, 1) & (1.5, 1) & (2.0, 1) & (3.0, 1)\\
          3  & $56\times56$   & 8 & 2 & MobileOne-Block & 128$\times\alpha$ & (1.0, 4)  & (1.5, 1) & (2.0, 1) & (2.5, 1) & (3.5, 1) \\
          4  & $28\times28$   & 5 & 2 & MobileOne-Block & 256$\times\alpha$ & (1.0, 4)  & (2.0, 1) & (2.5, 1) & (3.0, 1) & (3.5, 1) \\
          5  & $14\times14$   & 5 & 1 & MobileOne-Block & 256$\times\alpha$ & (1.0, 4)  & (2.0, 1) & (2.5, 1) & (3.0, 1) & (3.5, 1, SE-ReLU) \\
          6  & $14\times14$   & 1 & 2 & MobileOne-Block & 512$\times\alpha$ & (2.0, 4)  & (2.5, 1) & (4.0, 1) & (4.0, 1) & (4.0, 1, SE-ReLU) \\
          7  & $7\times7$     & 1 & 1 & AvgPool & - & - & - & - & - & - \\
          8  & $1\times1$     & 1 & 1 & Linear & 512$\times\alpha$ & 2.0 & 2.5 & 4.0 & 4.0 & 4.0 \\
          \midrule
        \end{tabular}
        \vspace{-0.2cm}
    \caption{MobileOne Network Specifications}
    \label{table:mobileone_network_specification}
    \vspace{-0.2cm}
\end{table*}

\paragraph{Model Scaling} 
Recent works scale model dimensions like width, depth, and resolution to improve performance \cite{efficientnet_v1_plmr,tinynet_neurips}. MobileOne has similar depth scaling as MobileNet-V2, i.e. using shallower early stages where input resolution is larger as these layers are significantly slower compared to later stages which operate on smaller input resolution. We introduce 5 different width scales as seen in Table~\ref{table:mobileone_network_specification}. Furthermore, we do not explore scaling up of input resolution as both FLOPs and memory consumption increase, which is detrimental to runtime performance on a mobile device. As our model does not have a multi-branched architecture at inference, it does not incur data movement costs as discussed in previous sections. This enables us to aggressively scale model parameters compared to competing multi-branched architectures like MobileNet-V2, EfficientNets, etc. without incurring significant latency cost. The increased parameter count enables our models to generalize well to other computer vision tasks like object detection and semantic segmentation (see Section~\ref{section:object_detection_mscoco}). In Table~\ref{tab:compare_traintime_overparam}, we compare against recent train time over-parameterization works ~\cite{Ding_2021_repvgg,ding2021diverse,Ding_2019_ICCV,NEURIPS2020_expandnets} and show that MobileOne-S1 variant outperforms RepVGG-B0 which is $\sim$3$\times$ bigger.

\subsection{Training}\label{sec:progressive_training}
As opposed to large models, small models need less regularization to combat overfitting. It is important to have weight decay in early stages of training as demonstrated empirically by~\cite{time_matters_weightdecay}. Instead of completely removing weight decay regularization as studied in~\cite{time_matters_weightdecay}, we find that annealing the loss incurred by weight decay regularization over the course of training is more effective. In all our experiments, we use cosine schedule~\cite{cosinelr_iclr2017} for learning rate. Further, we use the same schedule to anneal weight decay coefficient. We also use the progressive learning curriculum introduced in~\cite{efficientnet_v2_quoc}. 
In Table~\ref{table:pl_ablation}, we ablate over the various train settings keeping all other parameters fixed. We see that annealing the weight decay coefficient gives a 0.5\% improvement.
\begin{table}
  \footnotesize
  \centering
  \scalebox{0.9}{
  \begin{tabular}{lcccc}
        \toprule
        & \multirow{2}{*}{\bf Baseline} & {\bf + Progressive }  & {\bf + Annealing} & \multirow{2}{*}{\bf + EMA}\\
        & \textbf{} & {\bf Learning}  & {\bf Weight Decay} & \\
        \midrule
         \textbf{Top-1} & 76.4 & 76.8 & 77.3  & 77.4 \\
        \bottomrule
    \end{tabular}
    }
  \caption{Ablation on various train settings for MobileOne-S2 showing Top-1 accuracy on ImageNet.} \label{table:pl_ablation}
  \vspace{-0.2cm}
\end{table}

\subsection{Benchmarking}\label{sec:benchmarking}
Getting accurate latency measurements on a mobile device can be difficult. On the iPhone 12, there is no command line access or functionality to reserve all of a compute fabric for just the model execution. We also do not have access to the breakdown of the round-trip-latency into categories like the network initialization, data movement, and network execution. To measure latency, we developed an iOS application using swift \cite{swift_language}.
The application runs the models using Core ML \cite{coremltools}. To eliminate  startup  inconsistencies, the model graph is loaded, the input tensor is preallocated, and the model is run once before benchmarking begins. During benchmarking, the app runs the model many times (default is 1000) and statistic are accumulated. To achieve lowest latency and highest consistency, all other applications on the phone are closed. 
For the models latency seen in Table \ref{tab:compare_imagenet}, we report the full round-trip latency. A large fraction of this time may be from platform processes that are not model execution, but in a real application these delays may be unavoidable. Therefore we chose to include them in the reported latency.  In order to filter out interrupts from other processes, we report the minimum latency for all the models.
For CPU latency, we run the models on an Ubuntu desktop with a 2.3 GHz -- Intel Xeon Gold 5118 processor. For GPU latency, we compile the models using NVIDIA TensorRT library (v8.0.1.6) and run on a single RTX-2080Ti GPU with batch size set to 1. We report the median latency value out of 100 runs.

\section{Experiments}
\label{sec:experiments}
\paragraph{Image Classification on ImageNet-1K}\label{section:experiment_details_imagenet}
We evaluate MobileOne models on ImageNet~\cite{deng2009imagenet} dataset, which consists of 1.28 million training images and a validation set with 50,000 images from 1,000 classes. All models are trained from scratch using PyTorch~\cite{NEURIPS2019_9015} library on a machine with 8 NVIDIA GPUs. All models are trained for 300 epochs with an effective batch size of 256 using SGD with momentum~\cite{sgdsutskever13} optimizer. We use label smoothing regularization~\cite{Szegedy2016cv} with cross entropy loss with smoothing factor set to 0.1 for all models. The initial learning rate is 0.1 and annealed using a cosine schedule~\cite{cosinelr_iclr2017}. Initial weight decay coefficient is set to $10^{-4}$ and annealed to $10^{-5}$ using the same cosine schedule as described in~\cite{cosinelr_iclr2017}. We use AutoAugment~\cite{autoaugment_cvpr2019} to train only the bigger variants of MobileOne, i.e. S2, S3, and S4. The strength of autoaugmentation and image resolution is progressively increased during training as introduced in~\cite{efficientnet_v2_quoc}. We list the details in supplementary material. For smaller variants of MobileOne, i.e. S0 and S1 we use standard augmentation -- random resized cropping and horizontal flipping. We also use EMA (Exponential Moving Average) weight averaging with decay constant of 0.9995 for training all versions of MobileOne. 
At test time, all MobileOne models are evaluated on images of resolution $224\times224$. In Table~\ref{tab:compare_imagenet}, we compare against all recent efficient models that are evaluated on images of resolution $224\times224$ while having a parameter count $<$20 Million and trained without distillation as done in prior works like~\cite{mobilevit_iclr2022,mobileformer_cvpr2022}. FLOP counts are reported using the fvcore~\cite{fvcore} library. 

We show that even the smallest variants of transformer architectures have a latency upwards of 4ms on mobile device. Current state-of-the-art MobileFormer~\cite{mobileformer_cvpr2022} attains top-1 accuracy of 79.3\% with a latency of 70.76ms, while MobileOne-S4 attains 79.4\% with a latency of only 1.86ms which is $\sim$38$\times$ faster on mobile. MobileOne-S3 has 1\% better top-1 accuracy than EfficientNet-B0 and is faster by 11\% on mobile. Our models have a lower latency even on CPU and GPU compared to competing methods. 

\begin{table}[]
    \footnotesize
    \resizebox{\columnwidth}{!}{
    \begin{tabular}{lcccccc}
        \toprule
    \multirow{2}{*}{\textbf{Model}} & \multirow{2}{*}{\textbf{Top-1}} & {\textbf{FLOPs}} & {\textbf{Params}} & \multicolumn{3}{c}{\bf Latency (ms)} \\
    \cmidrule{5-7}
        & & \textbf{(M)} & \textbf{(M)} & \textbf{CPU} & \textbf{GPU} &\textbf{Mobile} \\
        \midrule
        \multicolumn{7}{c}{\textbf{Transformer Architectures}} \\ 
        \midrule
Mobileformer-96~\cite{mobileformer_cvpr2022} & 72.8 & 96 & 4.6 & 37.36 & - & 16.95 \\
ConViT-tiny~\cite{d2021convit} & 73.1 & 1000 & 5.7 & 28.95 & - & 10.99 \\
MobileViT-S~\cite{mobilevit_iclr2022} & 78.4 & 1792 & 5.6 & 30.76 &- & 9.21 \\
Mobileformer-52~\cite{mobileformer_cvpr2022} & 68.7 & 52 & 3.6 & 29.23 &- & 9.02 \\
PiT-ti~\cite{heo2021pit} & 71.3 & 710 & 4.9 & 16.37 & 1.97 & 8.81 \\
MobileViT-XS~\cite{mobilevit_iclr2022} & 74.8 & 941 & 2.3 & 27.21 &- & 6.97 \\
DeiT-tiny~\cite{Touvron_2021_ICCV} & 72.2 & 1300 & 5.9 & 16.68 & 1.78 & 4.78 \\
MobileViT-XXS~\cite{mobilevit_iclr2022} & 69.0 & 373 & 1.3 & 23.03  & -&  4.70 \\
        \midrule
        \multicolumn{7}{c}{\textbf{Convolutional Architectures}} \\ 
        \midrule
RepVGG-B1~\cite{Ding_2021_repvgg} & 78.4 & 11800 & 51.8 & 193.7 & 3.17 & 3.73 \\
RepVGG-A2~\cite{Ding_2021_repvgg} & 76.5 & 5100 & 25.5 & 93.43 & 2.41 & 2.41 \\
\textbf{MobileOne-S4} & 79.4 & 2978 & 14.8 & 26.60 & \bf 0.95 & \bf 1.86 \\
\midrule
RepVGG-B0~\cite{Ding_2021_repvgg} & 75.1 & 3100 & 14.3 & 55.97 & 1.45 & 1.82 \\
EfficientNet-B0~\cite{efficientnet_v1_plmr} & 77.1 & 390 & 5.3 & 28.71 & 1.35 &  1.72 \\
RepVGG-A1~\cite{Ding_2021_repvgg} & 74.5 & 2400 & 12.8 & 47.15 & 1.42 & 1.68\\
\textbf{MobileOne-S3} & 78.1 & 1896 & 10.1 & 16.47 & \bf 0.76 & \bf 1.53 \\
\midrule
MobileNetV2-x1.4~\cite{sandler2018mobilenetv2} & 74.7 & 585 & 6.9 & 15.67 & 0.80 & 1.36 \\
RepVGG-A0~\cite{Ding_2021_repvgg} & 72.4 & 1400 & 8.3 & 43.61 & 1.23 & 1.28 \\
MobileNeXt-x1.4~\cite{mobilenext_eccv2020} & 76.1 & 590 & 6.1 & 18.06 & 1.04 & 1.27 \\
\textbf{MobileOne-S2} & 77.4 & 1299 & 7.8 & 14.87 & \bf 0.72 & \bf 1.18 \\
\midrule
MixNet-S~\cite{mixconv_bmvc_2019} & 75.8 & 256 & 4.1 & 40.09 & 2.41 & 1.13 \\
MobileNetV3-L~\cite{Howard2019MobilenetV3} & 75.2 & 219 & 5.4 & 17.09 & 3.8 & 1.09 \\
ShuffleNetV2-2.0~\cite{Ma_2018_shufflenetv2} & 74.9 & 591 & 7.4 & 20.85 & 4.76 & 1.08 \\
MNASNet-A1~\cite{mnasnet_cvpr} & 75.2 & 312 & 3.9 & 24.06 & 0.95 & 1.00 \\
MobileNetV2-x1.0~\cite{sandler2018mobilenetv2} & 72.0 & 300 & 3.4 & 13.65 & 0.69 & 0.98 \\
MobileNetV1~\cite{Howard2017MobileNets} & 70.6 & 575 & 4.2 & 10.65 & 0.58 & 0.95 \\
MobileNeXt-x1.0~\cite{mobilenext_eccv2020} & 74.0 & 311 & 3.4 & 16.04 & 1.02 & 0.92 \\
\textbf{MobileOne-S1} & 75.9 & 825 & 4.8 & 13.04 & \bf 0.66 & \bf 0.89 \\
\midrule
MobileNetV3-S~\cite{Howard2019MobilenetV3} & 67.4 & 56 & 2.5 & 10.38 & 3.74 & 0.83 \\
ShuffleNetV2-1.0~\cite{Ma_2018_shufflenetv2} & 69.4 & 146 & 2.3 & 16.60 & 4.58 & 0.68 \\
\textbf{MobileOne-S0} & 71.4 & 275 & 2.1 & 10.55 & \bf 0.56 &\bf 0.79 \\
   
\bottomrule
    \end{tabular}}
\vspace{-0.2cm}
    \caption{Performance of various models on ImageNet-1k validation set. Note: All results are without distillation for a fair comparison. Results are grouped based on latency on mobile device. Models which could not be reliably exported either by TensorRT or Core ML Tools are annotated by ``-".}
    \label{tab:compare_imagenet}
\end{table} 

\vspace{-0.3cm}
\paragraph{Knowledge distillation}\label{section:kd}
Efficient models are often distilled from a bigger teacher model to further boost the performance. We demonstrate the performance of MobileOne backbones using state-of-the-art distillation recipe suggested in~\cite{shen2020mealv2}. From Table~\ref{tab:kd_inet}, our models outperform competing models of similar or higher parameter count. Train-time overparameterization enables our models to distill to better performance even though they have similar or smaller parameter count than competing models. In fact, MobileOne-S4 outperforms even ResNet-50 model which has 72.9\% more parameters. MobileOne-S0 has 0.4M less parameters at inference than MobileNetV3-Small and obtains 2.8\% better top-1 accuracy on ImageNet-1k dataset. 

\begin{table}[]
    \footnotesize
    \resizebox{\columnwidth}{!}{
    \begin{tabular}{lcccc}
        \toprule
    \multirow{2}{*}{\textbf{Model}} & {\textbf{Params}} & {\textbf{Latency}} & \multicolumn{2}{c}{\bf Top-1 Accuracy} \\
    \cmidrule{4-5}
         & \textbf{(M)} & \textbf{(ms)} & \textbf{Baseline} & \textbf{ Distillation} \\
        \midrule
        MobileNet V3-Small x1.0 & 2.5 & 0.83 & 67.4 & 69.7 \\
        \textbf{MobileOne-S0} & 2.1 & 0.79 & 71.4 & \textbf{72.5} \\
        
        \midrule
        MobileNet V3-Large 1.0 & 5.5 & 1.09 & 75.2 & 76.9 \\
        \textbf{MobileOne-S1} & 4.8 & 0.89 & 75.9 & \textbf{77.4} \\
        
        \midrule
        EfficientNet-B0 & 5.3 & 1.72 & 77.1 & 78.3 \\
        \textbf{MobileOne-S2} & 7.8 & 1.18 & 77.4 & \textbf{79.1} \\
        
        \midrule
        ResNet-18 & 11.7 & 2.10 & 69.8 & 73.2 \\
        \textbf{MobileOne-S3} & 10.1 & 1.53 & 78.1 & \textbf{80.0} \\
        
        \midrule
        ResNet-50 & 25.6 & 2.69 & 79.0 & 81.0 \\
        \textbf{MobileOne-S4} & 14.8 & 1.86 & 79.4 & \textbf{81.4} \\
   
\bottomrule
    \end{tabular}}
\vspace{-0.2cm}
    \caption{Performance of various models on ImageNet-1k validation set using MEAL-V2~\cite{shen2020mealv2} distillation recipe. Results of competing models are reported from~\cite{shen2020mealv2}. Models grouped based on parameter count.}
    \label{tab:kd_inet}
\end{table}

\vspace{-0.3cm}
\paragraph{Object detection on MS-COCO}\label{section:object_detection_mscoco}

To demonstrate the versatility of MobileOne, we use it as the backbone feature extractor for a single shot object detector SSD~\cite{Liu_ssd_2016}. Following~\cite{sandler2018mobilenetv2}, we replace standard convolutions in SSD head with separable convolutions, resulting in a version of SSD called SSDLite. The model is trained using the mmdetection library~\cite{mmdetection} on the MS COCO dataset~\cite{mscoco_lin_2014}. The input resolution is set to $320\times320$ and the model is trained for 200 epochs as described in~\cite{mobilevit_iclr2022}. For more detailed hyperparameters please refer to the supplementary material. We report mAP@IoU of 0.50:0.05:0.95 on the validation set of MS COCO in Table~\ref{tab:detection_segmentation}. Our best model outperforms MNASNet by 27.8\% and best version of MobileViT~\cite{mobilevit_iclr2022} by 6.1\%. We show qualitative results in the supplementary material.

\begin{table}[t]
       \resizebox{\columnwidth}{!}{
        \scriptsize
        \subfloat[]{
        \begin{tabular}{lc}
            \toprule
            \textbf{Feature backbone} & \textbf{mAP} ($\uparrow$)\\
            \midrule
            MobileNetV3~\cite{Howard2019MobilenetV3}    & 22.0 \\
            MobileNetV2~\cite{sandler2018mobilenetv2}    & 22.1 \\
            MobileNetV1~\cite{Howard2017MobileNets}    & 22.2 \\
            MixNet~\cite{mixconv_bmvc_2019}         & 22.3 \\
            MNASNet-A1~\cite{mnasnet_cvpr}        & 23.0 \\
            MobileVit-XS~\cite{mobilevit_iclr2022}  & 24.8 \\
            MobileViT-S~\cite{mobilevit_iclr2022}    & 27.7 \\
            \midrule
            \textbf{MobileOne-S1}  & {25.7} \\
            \textbf{MobileOne-S2}  & {26.6} \\
            \textbf{MobileOne-S3}  & {27.3} \\
            \textbf{MobileOne-S4}  & {29.4} \\
            \bottomrule
        \end{tabular}}
        \quad \quad
        \subfloat[]{
        \begin{tabular}{lcc}
            \toprule
            \multirow{2}{*}{\textbf{Feature backbone}} & 
            \multicolumn{2}{c}{\textbf{mIoU} ($\uparrow$)} \\
            \cmidrule{2-3}
            & \textbf{VOC} & \textbf{ADE20k} \\
            \midrule
            MobileNetV2-x0.5  &  70.2 & -\\
            MobileNetV2-x1.0  & 75.7 & 34.1\\
            MobileViT-XXS & 73.6 & -\\
            MobileViT-XS & 77.1 & -\\
            MobileViT-S  & 79.1 & -\\

            \midrule
            \textbf{MobileOne-S0}  & 73.7 & 33.1\\
            \textbf{MobileOne-S1}  & 77.3 & 35.1\\
            \textbf{MobileOne-S2}  & 77.9 & 35.7\\
            \textbf{MobileOne-S3}  & 78.8 & 36.2\\
            \textbf{MobileOne-S4}$^\dagger$  & {80.1} & {38.2}\\
            \bottomrule
        \end{tabular}}}
    \caption{(a) Quantitative performance of object detection on MS-COCO. (b) Quantitative performance of semantic segmentation on Pascal-VOC and ADE20k datasets. $^\dagger$This model was trained without Squeeze-Excite layers. } \label{tab:detection_segmentation}
\end{table} 

\vspace{-0.3cm}
\paragraph{Semantic Segmentation on Pascal VOC and ADE 20k}\label{section:segmentation}
We use MobileOne as the backbone for a Deeplab V3 segmentation network~\cite{deeplabv3} using the cvnets library \cite{mobilevit_iclr2022}. The VOC models were trained on the augmented Pascal VOC dataset ~\cite{pascal-voc-2012} for 50 epochs following the training procedure of \cite{mobilevit_iclr2022}. The ADE 20k\cite{ade20k} models were trained using the same hyperparameters and augmentations. For more detailed hyperparameters, please refer to the supplementary material. We report mean intersection-over-union (mIOU) results in Table~\ref{tab:detection_segmentation}. For VOC, our model outperforms Mobile ViT by $1.3\%$ and MobileNetV2 by $5.8\%$. Using the MobileOne-S1 backbone with a lower latency than the MobileNetV2-1.0 backbone, we still outperform it by $2.1\%$. For ADE 20k, our best variant outperforms MobileNetV2 by $12.0\%$. Using the smaller MobileOne-S1 backbone, we still outperform it by $2.9\%$. We show qualitative results in the supplementary material.

\vspace{-0.3cm}
\paragraph{Robustness to corruption}\label{section:robustness}
We evaluate MobileOne and competing models on the following benchmarks, ImageNet-A~\cite{imageneta}, a dataset that contains naturally occuring examples that are misclassified by resnets. ImageNet-R~\cite{imagenetr}, a dataset that contains natural renditions of ImageNet object classes with different textures and local image statistics. ImageNet-Sketch~\cite{imagenetsketch}, a dataset that contains black and white sketches of all ImageNet classes, obtained using google image queries. ImageNet-C~\cite{imagenetc}, a dataset that consists of algorithmically generated corruptions (blur, noise) applied to the ImageNet test-set. We follow the protocol set by~\cite{mao2022robust} for all the evaluations. We use pretrained weights provided by Timm Library~\cite{rw2019timm} for the evaluations. From Table~\ref{tab:robustness}, MobileOne outperforms other efficient architectures significantly on out-of-distribution benchmarks like ImageNet-R and ImageNet-Sketch. Our model is less robust to corruption when compared to MobileNetV3-L, but outperforms MobileNetV3-L on out-of-distribution benchmarks. Our model outperforms MobileNetV3-S, MobileNetV2 variants and EfficientNet-B0 on both corruption and out-of-distribution benchmarks as seen in Table~\ref{tab:robustness}.

\begin{table}[t]
\centering
\resizebox{\columnwidth}{!}{
\begin{tabular}{l c c c c c c c}
\toprule
\textbf{Model}  & \textbf{Latency(ms)} & \textbf{Clean} & \textbf{IN-C ($\downarrow$)} & \textbf{IN-A} & \textbf{IN-R} & \textbf{IN-SK} \\
\midrule
MobileNetV3-S & 0.83 & 67.9 & 86.5 & 2.0 & 27.3 & 16.2  \\
\textbf{MobileOne-S0} & 0.79 & \textbf{71.4} & \textbf{86.4} & \textbf{2.3} & \textbf{32.9} & \textbf{19.3} \\
\midrule
MixNet-S & 1.13 & 75.7 & 77.7 & \textbf{3.8} & 32.2 & 20.5 \\
MobileNetV3-L & 1.09 & 75.6 & \textbf{77.1} & 3.5 & 33.9 & \underline{22.6} \\
MobileNetV2-x1.0 & 0.98 & 73.0 & 84.1 & 2.1 & 32.5 & 20.8 \\
\textbf{MobileOne-S1} & 0.89 & \textbf{75.9} & 80.4 & 2.7 & \textbf{36.7} & \textbf{22.6} \\
\midrule
MobileNetV2-x1.4 & 1.36 & 76.5 & 78.9 & 3.7 & 36.0 & 23.7 \\
\textbf{MobileOne-S2} & 1.18 & \textbf{77.4} & \textbf{73.6} & \textbf{4.8} & \textbf{40.0} & \textbf{26.4} \\
\midrule
EfficientNet-B0 & 1.72 & 77.6 & 72.2 & 7.2 & 36.6 & 25.0 \\
\textbf{MobileOne-S3} & 1.53 & 78.1 & 71.6 & 7.1 & \textbf{42.1} & 28.5 \\
\textbf{MobileOne-S4} & 1.86 & \textbf{79.4} & \textbf{68.1} & \textbf{10.8} & 41.8 & \textbf{29.2} \\
\bottomrule

\end{tabular}
}
\caption{Results on robustness benchmark datasets following protocol set by~\cite{mao2022robust}. For ImageNet-C mean corruption error is reported (lower is better) and for other datasets Top-1 accuracy is reported (higher is better). Results are grouped following Table~\ref{tab:compare_imagenet}}
\label{tab:robustness}
\vspace{-0.2cm}
\end{table}

\vspace{-0.3cm}
\paragraph{Comparison with Micro Architectures}\label{section:micro_arch}
Recently~\cite{Li_2021_ICCV_micronet, tinynet_neurips} introduced architectures that were extremely efficient in terms of FLOPS and parameter count. But architectural choices introduced in these micro architectures like~\cite{Li_2021_ICCV_micronet}, do not always result in lower latency models. MicroNet uses dynamic activations which are extremely inefficient as demonstrated in Table~\ref{tab:activation_latency}. In fact, smaller variants of MobileOne can easily outperform previous state-of-the-art micro architectures. Please see supplementary materials for more details on MobileOne micro architectures. In Table~\ref{tab:compare_micro_imagenet}, our models have similar latency as TinyNets, but have significantly lower parameter count and better top-1 accuracy. MobileOne-$\mu$1, is 2$\times$ smaller and has 6.3\% better top-1 accuracy while having similar latency as TinyNet-E.

\begin{table}[]
    \centering
    \resizebox{0.9\columnwidth}{!}{
    \begin{tabular}{lcccc}
        \toprule
    \multirow{2}{*}{\textbf{Model}} & \multirow{2}{*}{\textbf{Top-1}} & {\textbf{FLOPs}} & {\textbf{Params}} & {\bf Mobile } \\
        & & \textbf{(M)} & \textbf{(M)} &\textbf{Latency (ms)} \\
    \midrule
    TinyNet-D~\cite{tinynet_neurips} & 67.0 & 52 & 2.3 & 0.51 \\
    \bf MobileOne-$\boldsymbol \mu$2 & \bf 69.0 & 214 & 1.3 & \bf 0.50 \\ 
    \midrule
    MicroNet-M3~\cite{Li_2021_ICCV_micronet} & 62.5 & 20 & 2.6 & 12.02 \\
    MicroNet-M2~\cite{Li_2021_ICCV_micronet} & 59.4 & 12 & 2.4 & 9.49 \\
    TinyNet-E~\cite{tinynet_neurips} & 59.9 & 24 & 2.0 & 0.49 \\
    \bf MobileOne-$\boldsymbol \mu$1 & \bf 66.2 & 139 & 0.98 & \bf 0.47 \\ 
    \midrule
    MicroNet-M1~\cite{Li_2021_ICCV_micronet} & 51.4 & 6 & 1.8 & 3.33 \\
    \bf MobileOne-$\boldsymbol \mu$0 & \bf 58.5 & 68 & 0.57 & \bf 0.45 \\ 
    \bottomrule
    \end{tabular}}
\vspace{-0.2cm}
    \caption{Performance of various micro-architecture models on ImageNet-1k validation set. Note, we replace swish activations with ReLU in TinyNets for a fair comparison.}
    \label{tab:compare_micro_imagenet}
\end{table} 

\section{Discussion}
We have proposed an efficient, general-purpose backbone for mobile devices. Our backbone is suitable for general tasks such as image classification, object detection and semantic segmentation. We show that in the efficient regime, latency may not correlate well with other metrics like parameter count and FLOPs. Furthermore, we analyze the efficiency bottlenecks for various architectural components used in modern efficient CNNs by measuring their latency directly on a mobile device. We empirically show the improvement in optimization bottlenecks with the use of re-parameterizable structures. Our model scaling strategy with the use of re-parameterizable structures attains state-of-the-art performance while being efficient both on a mobile device and a desktop CPU.

\vspace{-0.3cm}
\paragraph{Limitations and Future Work} Although, our models are state-of-the-art within the regime of efficient architectures, the accuracy lags large models~\cite{liu2022convnext, liu2021swin}.
Future work will aim at improving the accuracy of these lightweight models. 
We will also explore the use of our backbone for faster inference on other computer vision applications not explored in this work such as optical flow, depth estimation, 3D reconstruction, etc.

{\small
\bibliographystyle{ieee_fullname}
\bibliography{bibliography}
}

\appendix
\section{Figures}
Figure 1 from the main paper has been enlarged in Figures~\ref{fig:top1_vs_latency_zoomed}, ~\ref{fig:top1_vs_latency}, ~\ref{fig:top1_vs_map}.
\begin{figure*}[!htb]
    \centering
    \includegraphics[width=0.95\linewidth]{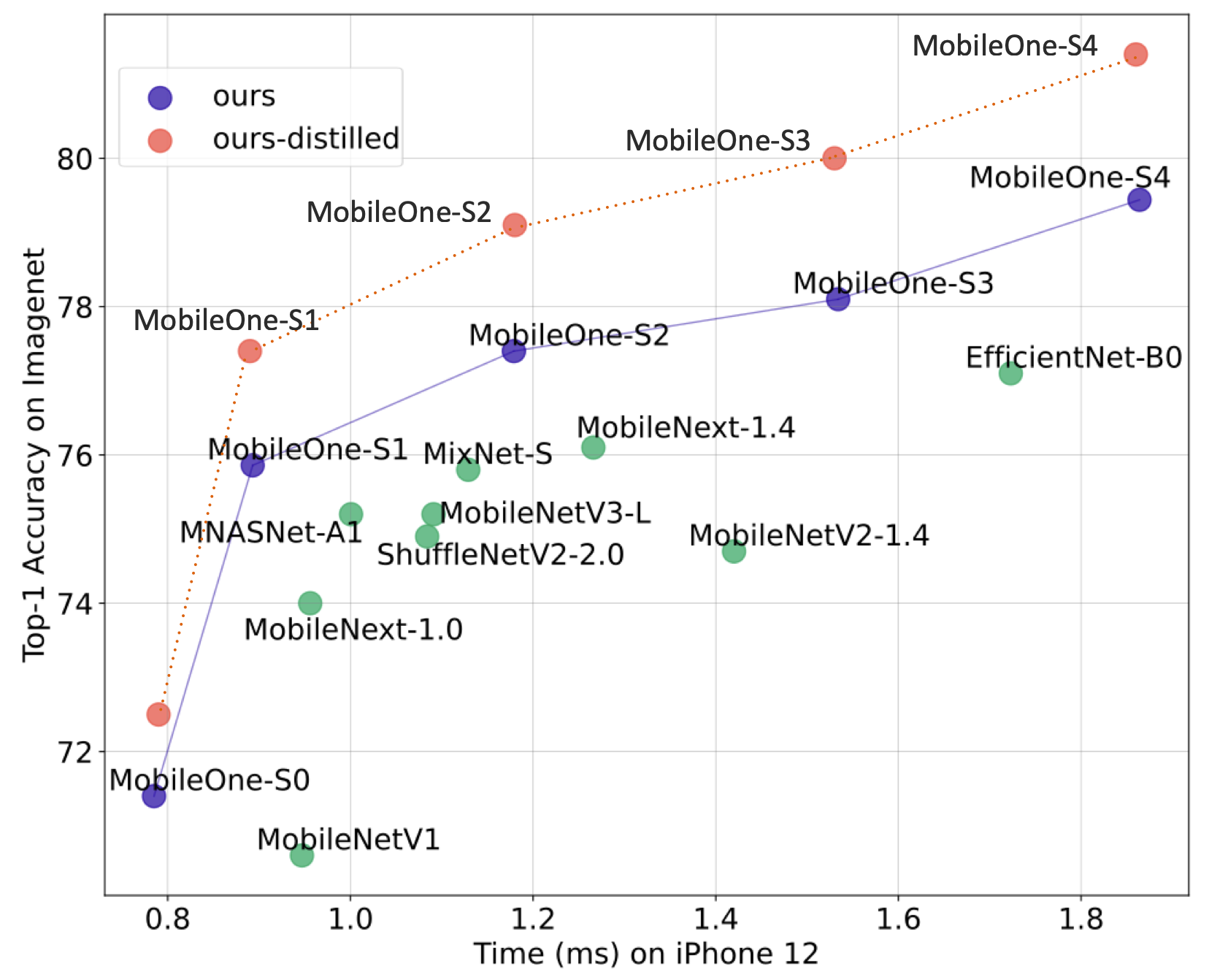}%
    \caption
      {%
            Top 1 accuracy vs Latency on iPhone 12. Corresponds to Figure 1a in the main paper.%
            \label{fig:top1_vs_latency_zoomed}
      }%
\end{figure*}

\begin{figure}[!htb]
  \centering
  \includegraphics[width=0.95\linewidth]{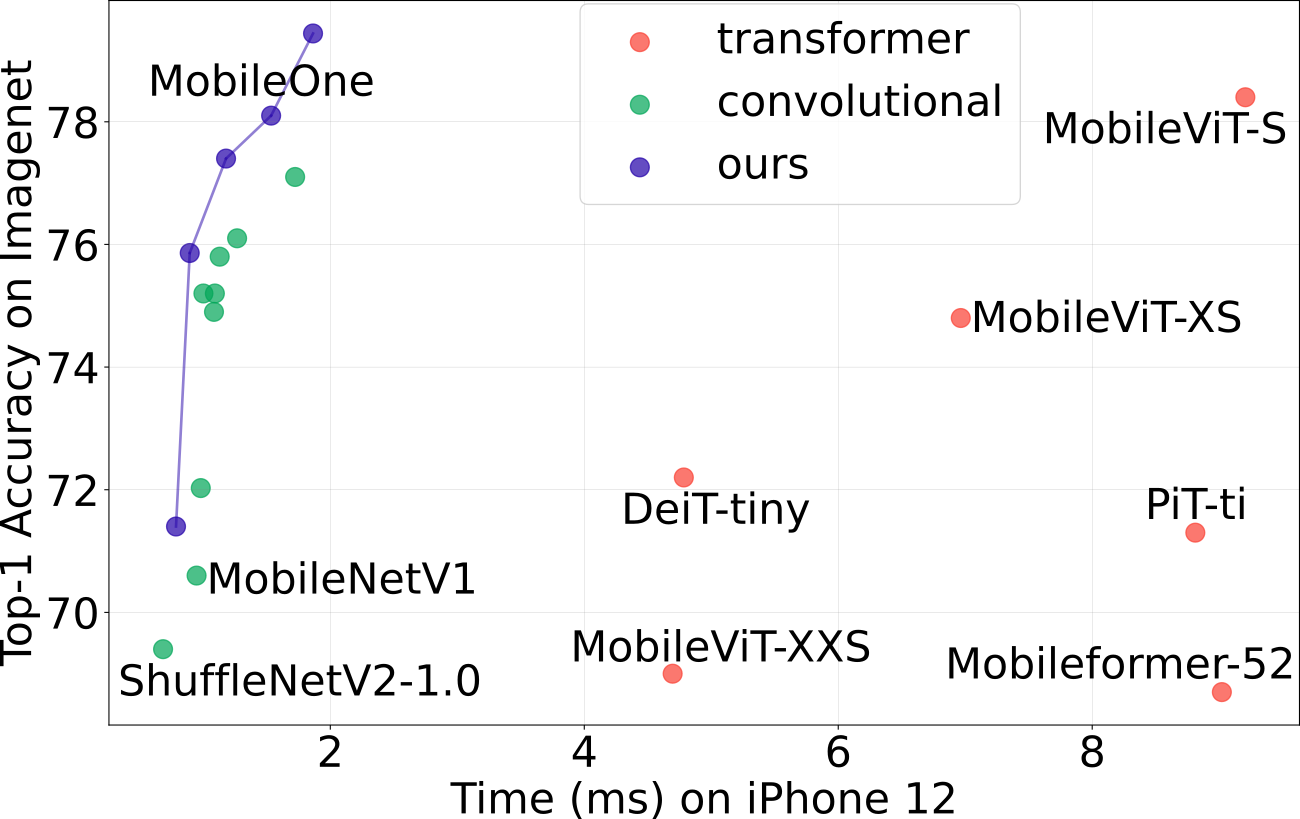}%
  \caption
    {%
      Zoomed out (a).  Corresponds to Figure 1b in the main paper.
      }
    \label{fig:top1_vs_latency}
               \vspace{0.5cm}

\end{figure}

\begin{figure}[!htb]
    \centering
  \includegraphics[width=0.95\linewidth,page=2]{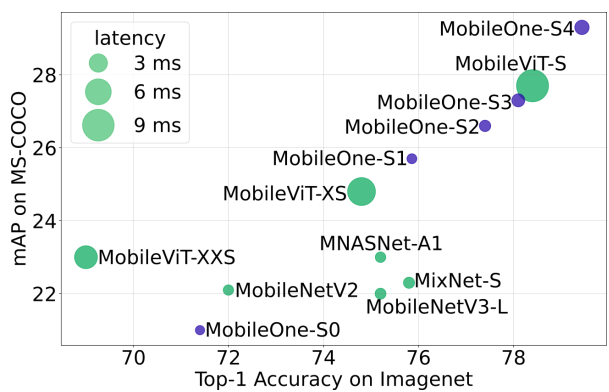}%
  \caption
    {%
      Top-1 accuracy vs mAP. Corresponds to Figure 1c in the main paper.
    }%
        \label{fig:top1_vs_map}
        \vspace{0.5cm}
\end{figure}

\section{Benchmarking}
We treat MobileNetV3~\cite{Howard2019MobilenetV3} in a special way since their H-swish operator is optimized for certain hardware platforms and not for others. Howard et al. \cite{Howard2019MobilenetV3} show that H-swish can obtain similar performance as ReLU when platform specific optimizations are applied. Therefore, while benchmarking for latency, we replace the H-swish layers with ReLU layers and then report the latency of MobileNetV3.

\subsection{Additional Benchmarks}
We have shown the efficiency of our model with comparisons on CPU, desktop GPU (RTX-2080Ti) and Mobile (iPhone 12). Additionally, in Table~\ref{tab:pixelphones}, we port existing architectures to Pixel-6 TPU and compare with our model. We observe that MobileOne achieves state-of-the-art accuracy-latency trade-off on TPU as well.

\begin{table}[t]
\scalebox{0.76}{
    \begin{tabular}{lccccc}
    \toprule
    &    & \multicolumn{4}{c}{\textbf{Latency (ms)} $\downarrow$ }            \\
    \textbf{Model}     &  Top1 $\uparrow$ & CPU   & iPhone12   & TensorRT          & Pixel-6$^\dagger$ \\
                      &       &  (x86)  & (ANE) & (2080Ti)         & (TPU) \\
    \midrule
    RepVGG-B2            & 78.8& 492.8 & 6.38     & 4.79             & 6.83 \\ 
    RepVGG-B1            &78.4& 193.7 & 3.73     & 3.17              & 4.28 \\
    RepVGG-A2            &76.5& 93.43 & 2.41     & 2.41              & 2.28 \\
    \textbf{MobileOne-S4}& 79.4& 26.6  & 1.86     & 0.95             & 2.17 \\
    \midrule
    EfficientNet-B0      & 77.1& 28.71 & 1.72     & 1.35             & 2.49 \\
    \textbf{MobileOne-S3}& 78.1& 16.47 & 1.53     & 0.76             & 1.28 \\
     \midrule
    RepVGG-B0            &75.1& 55.97 & 1.82     & 1.42              & 1.43 \\
    RepVGG-A1            & 74.5& 47.15 & 1.68     & 1.42              & 1.21   \\
    \textbf{MobileOne-S2}& 77.4& 14.87 & 1.18     & 0.72             & 1.07  \\
    \midrule
    RepVGG-A0            &72.4& 43.61 & 1.23     & 1.28              & 1.01   \\
    MobileNetV3-L        & 75.2& 17.09 & 1.09     & 3.8              & 1.01   \\
    MobileNetV2-x1.4     & 74.7& 15.67 & 1.36     & 0.8              & 0.98   \\
    MNASNet-A1           & 75.8& 24.06 & 1.00       & 0.95             & 0.88  \\
    MobileNetV2-x1.0     & 72.0& 13.65 & 0.98     & 0.69             & 0.77 \\
    \textbf{MobileOne-S1}& 75.9& 13.04 & 0.89     & 0.66              & 0.79 \\
    \midrule
    MobileNetV3-S        & 67.4& 10.38 & 0.83     & 3.74              & 0.67 \\
    ShuffleNetV2-x1.0    & 69.4& 16.6  & 0.68     & 4.58              & - \\
    MobileNetV1          & 70.6& 10.65 & 0.95     & 0.58             & 0.73 \\
    \textbf{MobileOne-S0}& 71.4& 10.55 & 0.79     & 0.56              & 0.59    \\
    \bottomrule
    \end{tabular}
    }
    \caption{Comparison with mobile architectures on Intel Xeon CPU, NVIDIA 2080Ti GPU, iPhone 12 and Pixel-6. ``$\dagger$" denotes models on Pixel-6 TPU, where weights and activations were converted to \texttt{int8} format. For all other compute platforms, models were evaluated in \texttt{fp16} format.}
\label{tab:pixelphones}
\end{table}

\begin{table}[!htb]
\centering
\scalebox{0.85}{
  \begin{tabular}{ccc}
    \toprule
    \multirow{ 2}{*}{\textbf{Epoch Range}} & \multirow{ 2}{*}{\textbf{Image Resolution}} & \textbf{AutoAugment} \\
     & & \textbf{Strength} \\
    \midrule
     0 - 38 & 160 & 0.3 \\
     39 - 113 & 192 & 0.6 \\
     114 - 300 & 224 & 1.0 \\
    \bottomrule
  \end{tabular}
  }
  \caption{Progressive training settings. AutoAugment is used only for training MobileOne-S2,S3,S4 variants.}\label{table:pl_curriculum}
\end{table}

\begin{table*}[]
    \scalebox{0.8}{
    \begin{tabular}{lccl}
    \toprule
        \multirow{2}{*}{Model}	& Top-1 & Mobile & \multirow{2}{*}{Training Recipe} \\
        & Accuracy & Latency(ms) &  \\
    \midrule \midrule
        MobileOne-S4 (Ours) &	79.4	& 1.86 &	CosLR + EMA + AA + PL + AWD \\
        MobileOne-S3 (Ours) &	78.1	& 1.53 &	CosLR + EMA + AA + PL + AWD \\
        \midrule
        EfficientNet-B0	& 77.1	& 1.72 &	Baseline reported by respective authors \\
        EfficientNet-B0	& 77.4	& 1.72 &	WCosLR + EMA + RA + RandE + DropPath + Dropout (Baseline reproduced) \\
        EfficientNet-B0	& 77.8	& 1.72 &	WCosLR + EMA + RA + RandE + DropPath + Dropout + PL + AWD \\
        EfficientNet-B0	& 74.9	& 1.72 &	CosLR + EMA + AA + PL + AWD \\
        \midrule \midrule
        MobileOne-S2 (Ours)	& 77.4	& 1.18	& CosLR + EMA + AA + PL + AWD \\
        \midrule
        MobileNetV2 $\times$1.4	& 74.7	& 1.36 & 	Baseline reported by respective authors \\
        MobileNetV2 $\times$1.4	& 75.7	& 1.36 & 	WCosLR + EMA + RA + RandE + DropPath + Dropout (Baseline reproduced) \\
        MobileNetV2 $\times$1.4	& 76.2	& 1.36	& WCosLR + EMA + RA + RandE + DropPath + Dropout + PL + AWD \\
        MobileNetV2 $\times$1.4	& 76.0	& 1.36	& CosLR + EMA + AA + PL + AWD \\
        \midrule \midrule
        MobileOne-S1 (Ours)	& 75.9	& 0.89	& CosLR + EMA + PL + AWD \\ \midrule
    
        MixNet-S	& 75.8	& 1.13 &	Baseline reported by respective authors \\
        MixNet-S	& 75.6	& 1.13 &	WCosLR + EMA + DropPath (Baseline reproduced) \\
        MixNet-S	& 75.4	& 1.13 &	WCosLR + EMA + DropPath + PL + AWD \\
        MixNet-S	& 75.5	& 1.13 &	CosLR + EMA + PL + AWD \\
        \midrule
        MobileNetV3-L	& 75.2	& 1.09 &	Baseline reported by respective authors \\
        MobileNetV3-L	& 75.4	& 1.09 &	WCosLR + EMA + RA + RandE + DropPath + Dropout + LR Noise (Baseline reproduced) \\
        MobileNetV3-L	& 75.6	& 1.09 &	WCosLR + EMA + RA + RandE + DropPath + Dropout + LR Noise + PL + AWD \\
        MobileNetV3-L	& 72.5	& 1.09	 & CosLR + EMA + AA + PL + AWD \\
        \midrule
        MobileNetV2 $\times$1.0	& 72.0 &	0.98	& Baseline reported by respective authors \\
        MobileNetV2 $\times$1.0	& 72.9 &	0.98	& WCosLR + EMA (Baseline reproduced) \\
        MobileNetV2 $\times$1.0	& 73.0	& 0.98	& WCosLR + EMA + PL + AWD \\
        \midrule
        MobileNetV1	& 70.6	& 0.95 &	Baseline reported by respective authors \\
        MobileNetV1	& 72.7	& 0.95 &	CosLR + EMA (Baseline reproduced) \\
        MobileNetV1	& 73.7	& 0.95 &	CosLR + EMA + PL + AWD \\
        \bottomrule
    \end{tabular}}
    \\
    \scalebox{0.8}{
    \begin{tabular}{ll}
    
    \toprule
    Legend \\
    \midrule
    AA & AutoAugment \\
    RA & RandAugment \\ 
    PL & Progressive Learning \\ 
    AWD & Annealing Weight Decay \\ 
    RandE & Random Erasing \\ 
    EMA & Exponential Moving Average \\ 
    CosLR & Cosine learning rate schedule \\
    WCosLR & Cosine learning rate schedule with Warmup \\
    LR Noise & Learning Rate Noise schedule in Timm \\
    \bottomrule
    \end{tabular}}
    
    \caption{Top-1 Accuracy on ImageNet-1k for various training recipes.}
    \label{tab:train_recipe_abl}
\end{table*}

\section{Image Classification}
\subsection{Training details}

All models are trained from scratch using PyTorch~\cite{NEURIPS2019_9015} library on a machine with 8 NVIDIA A100 GPUs. All models are trained for 300 epochs with an effective batch size of 256 using SGD with momentum~\cite{sgdsutskever13} optimizer. We follow progressive training curriculum~\cite{efficientnet_v2_quoc} for faster training and better generalization. Throughout training the image resolution and the augmentation strength($\alpha$) is gradually increased, see Table~\ref{table:pl_curriculum}. The magnitude for augmentations in AutoAugment~\cite{autoaugment_cvpr2019} policy are between 0-9, we simply multiply $\alpha$ with this value to simulate variable strength of autoaugmentation. AutoAugment~\cite{autoaugment_cvpr2019} is used to train only the bigger variants of MobileOne, i.e. S2, S3, and S4. For smaller variants of MobileOne, i.e. S0 and S1 we use standard augmentation -- random resized cropping and horizontal flipping.
We use label smoothing regularization~\cite{Szegedy2016cv} with cross entropy loss with smoothing factor set to 0.1 for all models. The initial learning rate is 0.1 and annealed using a cosine schedule~\cite{cosinelr_iclr2017}. Initial weight decay coefficient is set to $10^{-4}$ and annealed to $10^{-5}$ using the same cosine schedule. We also use EMA (Exponential Moving Average) weight averaging with decay constant of 0.9995 for training all versions of MobileOne.

\begin{table*}[]
\centering
\footnotesize
        \begin{tabular}{cccccccc}
          \toprule
          \multirow{2}{*}{\textbf{Stage}} & \multirow{2}{*}{\textbf{Input}} & \multirow{2}{*}{\textbf{Stride}} & \multirow{2}{*}{\textbf{Block Type}} &
          \multirow{2}{*}{\textbf{\# Channels}} & \multicolumn{3}{c}{\textbf{(\# Blocks, $\alpha$, $k$) act=ReLU}} \\
          \cmidrule{6-8}
          & & & & &\textbf{$\boldsymbol \mu$0} & \textbf{$\boldsymbol \mu$1} & \textbf{$\boldsymbol \mu$2}\\
          \midrule
          1  & $224\times224$ & 2 & MobileOne-Block & 64$\times\alpha$  & (1, 0.75, 3)  & (1, 0.75, 2) & (1, 0.75, 2) \\
          2  & $112\times112$ & 2 & MobileOne-Block & 64$\times\alpha$  & (2, 0.75, 3)  & (2, 0.75, 2) & (2, 0.75, 2) \\
          3  & $56\times56$   & 2 & MobileOne-Block & 128$\times\alpha$ & (4, 0.5,  3)  & (6, 0.75, 2) & (6, 1.0, 2) \\
          4  & $28\times28$   & 2 & MobileOne-Block & 256$\times\alpha$ & (3, 0.5,  3)  & (4, 0.75, 2) & (4, 1.0, 2)  \\
          5  & $14\times14$   & 1 & MobileOne-Block & 256$\times\alpha$ & (3, 0.5,  3)  & (4, 0.75, 2) & (4, 1.0, 2)  \\
          6  & $14\times14$   & 2 & MobileOne-Block & 512$\times\alpha$ & (1, 0.75, 3)  & (1, 1.0,  2) & (1, 1.0, 2)  \\
          7  & $7\times7$     & 1 & AvgPool & - & - & - & -  \\
          8  & $1\times1$     & 1 & Linear & 512$\times\alpha$ & 0.75 & 1.0 & 1.0  \\
          \midrule
        \end{tabular}
    \caption{MobileOne micro variant specifications. }
    \label{table:mobileone_micro}
\end{table*}

\subsection{Analysis of Training Recipes}
Recent models introduce their own training recipe including regularization techniques to train them to competitive accuracies. We ablate over some of the commonly used recipes to train EfficientNet, MobileNetV3-L, MixNet-S, MobileNetV2 and MobileNetV1 in Table~\ref{tab:train_recipe_abl}. Mainly, we report the following,
\begin{itemize}
\item Results from original training recipes of the respective models. (baselines)
\item Results from training the models using recipe used to train MobileOne models.
\item Results obtained by adding EMA, Progressive Learning (PL) and Annealing Weight decay (AWD) to the original recipe proposed by respective works.
\end{itemize}
All runs below have been reproduced using Timm library~\cite{rw2019timm}. For a fair comparison all models are trained for 300 epochs. From Table~\ref{tab:train_recipe_abl}, we observe that our models use less regularization techniques as opposed to competing models like EfficientNet, MobileNetV3-L and MixNet-S to reach competitive accuracies. When we apply our training recipe to the competing models, there is no improvement in models like EfficientNet, MobileNetV3-L and MixNet-S. There are slight improvements in MobileNetV2 and MobileNetV1. However, the accuracy at iso-latency gap between our models is still large. When progressive learning and annealing weight decay is used with baseline recipes, we obtain additional improvements, for example MobileNetV1, gets 1\% improvement and MobileNetV2 $\times$1.4 gets 0.5\% improvement. 

\subsection{Sensitivity to Random Seeds}
Our model and training runs are stable and give similar performance with different random seeds, see Table~\ref{table:sensitivity}.

\begin{table}[!htb]
  \centering
  \scalebox{0.85}{
  \begin{tabular}{ccc}
    \toprule
    \textbf{Model} & \textbf{Run \#1} & \textbf{Run \#2} \\
    \midrule
     MobileOne-S0 & 71.402 & 71.304 \\
     MobileOne-S1 & 75.858 & 75.877 \\
     MobileOne-S2 & 77.372 & 77.234 \\
     MobileOne-S3 & 78.082 & 78.008 \\
     MobileOne-S4 & 79.436 & 79.376 \\
    \bottomrule
  \end{tabular}
  }
  \caption{Runs from 2 different seeds for all variants of MobileOne}\label{table:sensitivity}
\end{table}

\section{Micro Architectures}
In Table~\ref{table:mobileone_micro}, we provide specifications for micro variants of MobileOne introduced in Table 13 of main paper. Rather than optimizing for FLOPs, as done in~\cite{Li_2021_ICCV_micronet, tinynets_neurips} we sample variants that are significantly smaller in parameter count and use trivial overparameterization to train these architectures to competitive accuracies. 

\subsection{Effectiveness of Overparameterization}
We find that additional overparameterization branches benefits smaller variants more than it does for larger variants. In our experiments, we found that smaller variants improve consistently with additional overparameterization branches. Note, for all the experiments in Table~\ref{tab:overparam}, we use the same hyperparameters as described in Section 4 of main paper.

\begin{figure}[t]
    \centering
    \includegraphics[width=\columnwidth]{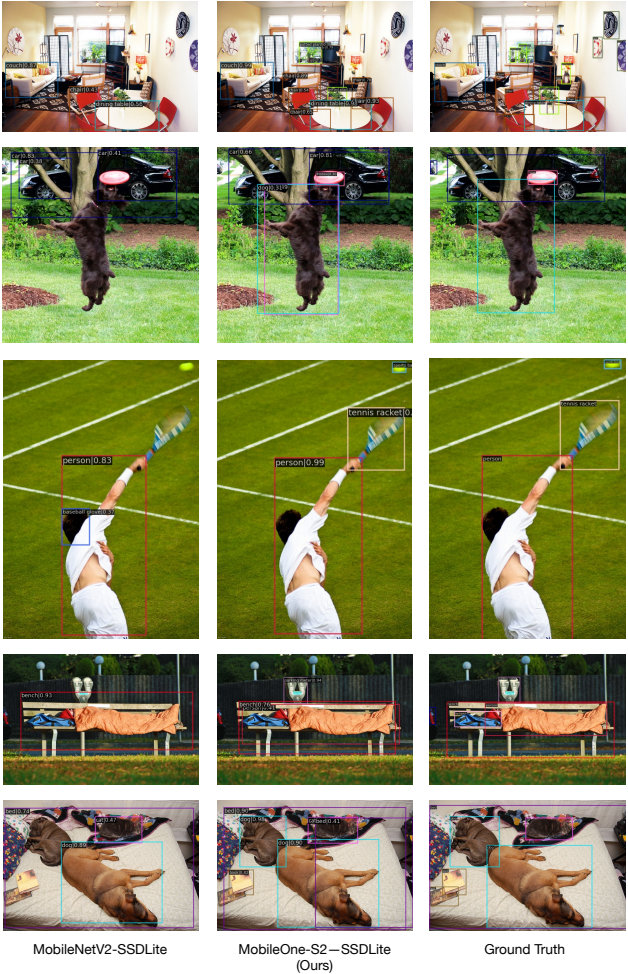}%
    \caption
      {%
            Qualitative comparison of MobileOne-S2-SSDLite (middle) against MobileNetV2-SSDLite (left) and ground truth (right). The two models have similar latency.%
            \label{fig:ssdlite_results}
      }%
\end{figure}
\begin{table}[]
    \centering
    \small
    \begin{tabular}{lccc}
    \toprule
         &  $k=1$ & $k=2$ & $k=3$ \\
     \midrule
       MobileOne-$\mu1$ & 65.7 & 66.2 & 65.9 \\
       MobileOne-$\mu2$ & 68.6 & 69.0 & 68.8 \\
       MobileOne-S0 & 70.9 & 70.7 & 71.3 \\
    \bottomrule
    \end{tabular}
    \caption{Effect of over-parametrization factor $k$ on MobileOne variants. Top-1 accuracy on ImageNet is reported.}
    \label{tab:overparam}
\end{table}
\section{Object Detection}
\subsection{Training details}
SSDLite models were trained for 200 epochs using cosine learning rate schedule with warmup, following~\cite{mobilevit_iclr2022}. Linear warmup schedule with a warmup ratio of 0.001 for 4500 iterations was used. Image size of 320$\times$320 was used for both training and evaluation, following~\cite{mobilevit_iclr2022}. We used SGD with momentum optimizer~\cite{sgdsutskever13} with an initial learning rate of 0.05, momentum of 0.9 and weight decay of 0.0001 for all the models. We use an effective batchsize of 192, following~\cite{mmdetection}. 
The models were trained on a machine with 8 NVIDIA A100 GPUs.

\subsection{Qualitative Results}
Visualizations in Figure~\ref{fig:ssdlite_results} are generated using \texttt{image\_demo.py}~\cite{mmdetection} with default thresholds in MMDetection library~\cite{mmdetection}. We compare MobileNetV2-SSDLite with MobileOne-S2-SSDLite which have similar latencies. Our model outperforms MobileNetV2-SSDLite in detecting small and large objects. In the first row, our model detects the potted plants amongst all the clutter in the scene. In the second row, our model detects both the dog and frisbee as opposed to MobileNetV2. In the third row, our model detects the tennis racket and the ball even though they are blurry. In the remaining rows, our model consistently detects both small and large foreground objects as opposed to MobileNetV2.

\begin{figure*}
    \centering
    \includegraphics[width=0.24\textwidth]{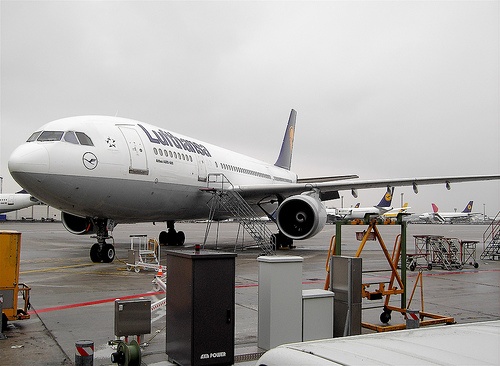}
    \includegraphics[width=0.24\textwidth]{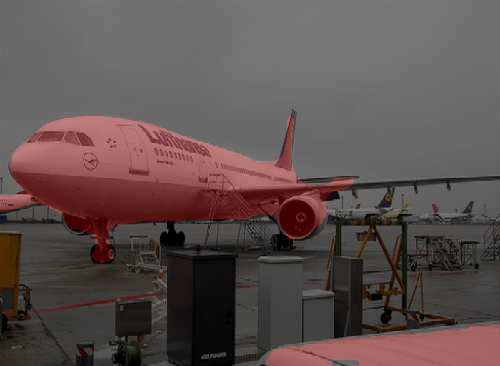}
    \includegraphics[width=0.24\textwidth]{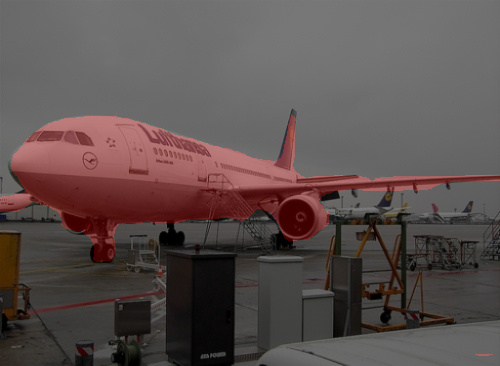}
    \includegraphics[width=0.24\textwidth]{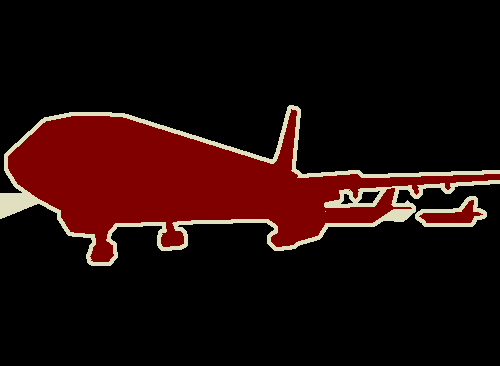}
 \\
    \includegraphics[width=0.24\textwidth]{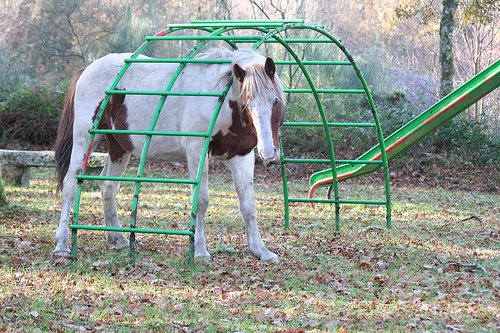}
    \includegraphics[width=0.24\textwidth]{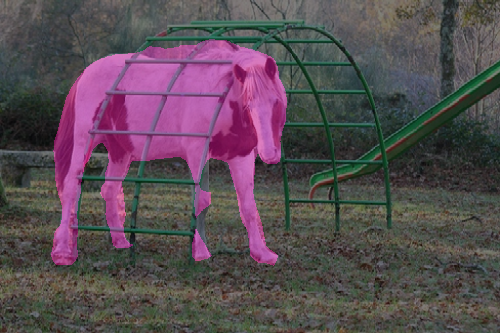}
    \includegraphics[width=0.24\textwidth]{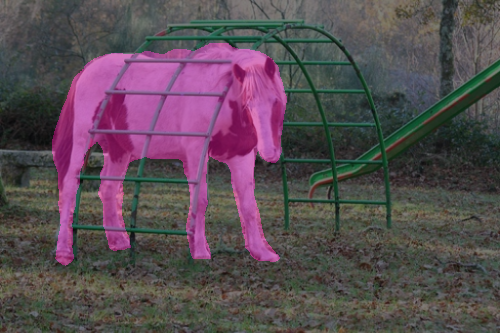}
    \includegraphics[width=0.24\textwidth]{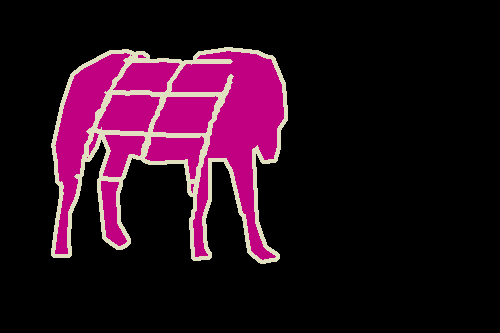}
\\
    \includegraphics[width=0.24\textwidth]{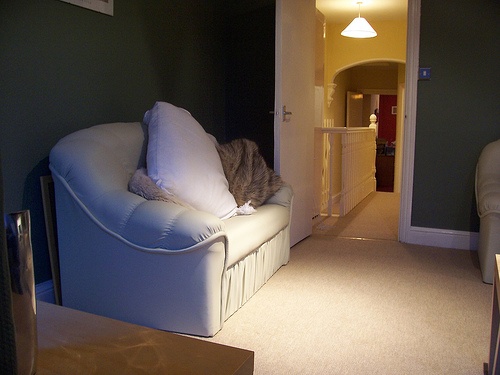}
    \includegraphics[width=0.24\textwidth]{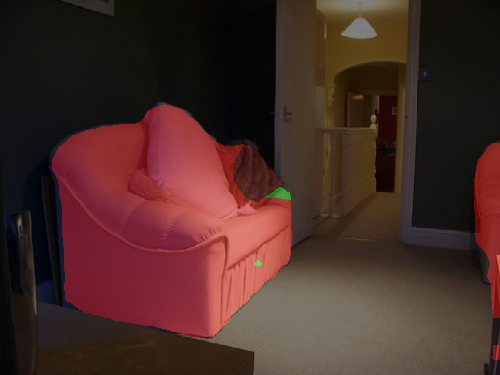}
    \includegraphics[width=0.24\textwidth]{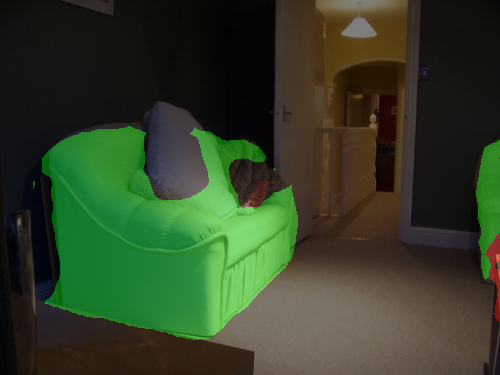}
    \includegraphics[width=0.24\textwidth]{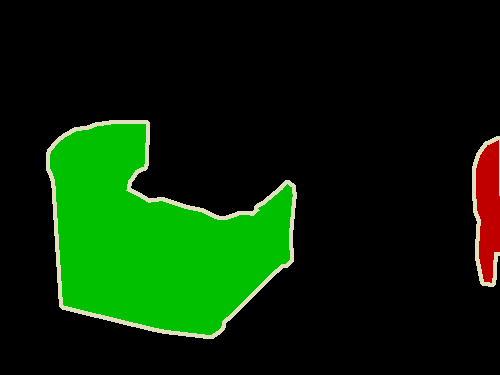}
\\
    \includegraphics[width=0.24\textwidth]{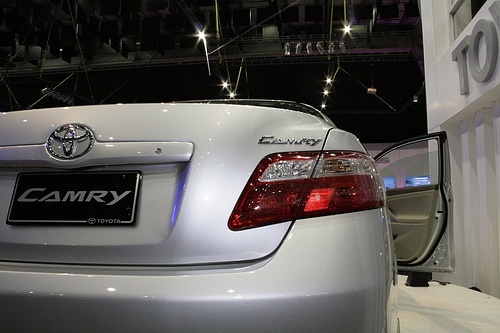}
    \includegraphics[width=0.24\textwidth]{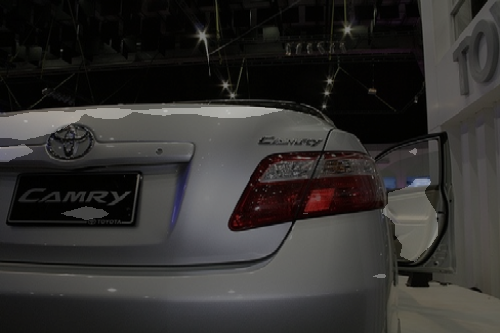}
    \includegraphics[width=0.24\textwidth]{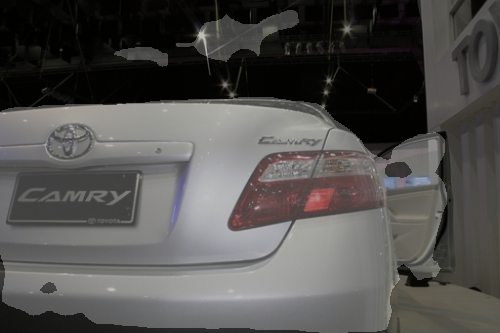}
    \includegraphics[width=0.24\textwidth]{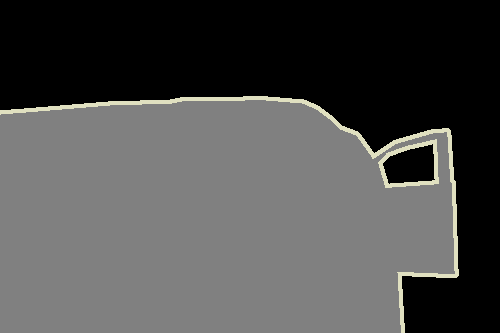}
\\
    \includegraphics[width=0.24\textwidth]{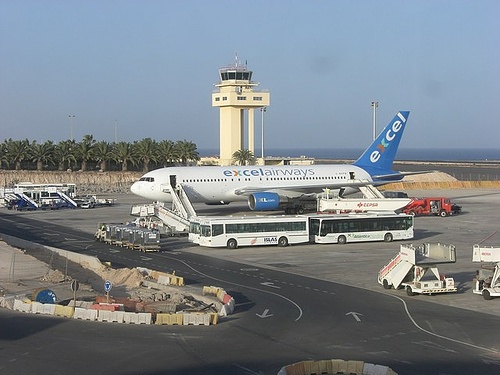}
    \includegraphics[width=0.24\textwidth]{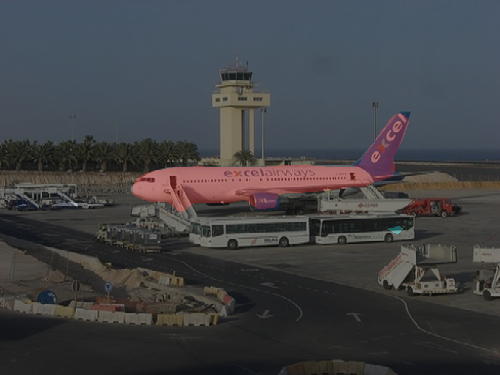}
    \includegraphics[width=0.24\textwidth]{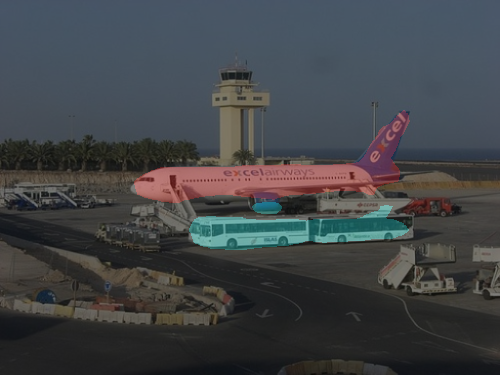}
    \includegraphics[width=0.24\textwidth]{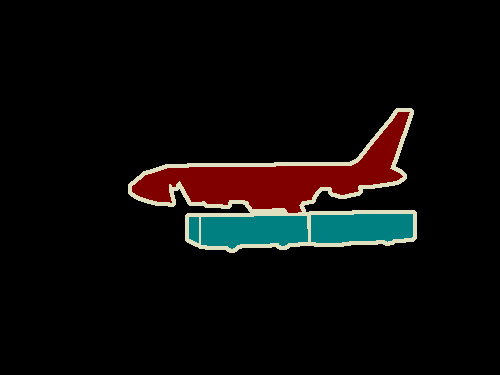}
\\
    \includegraphics[width=0.24\textwidth]{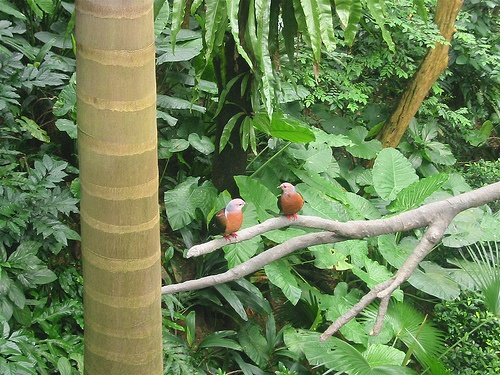}
    \includegraphics[width=0.24\textwidth]{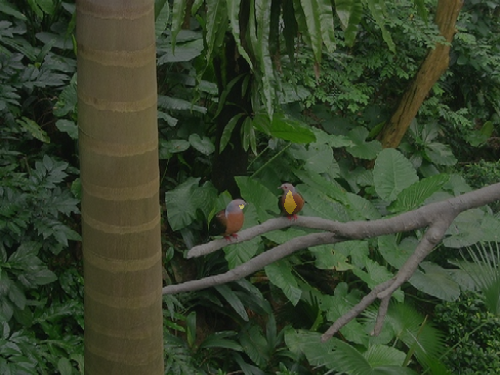}
    \includegraphics[width=0.24\textwidth]{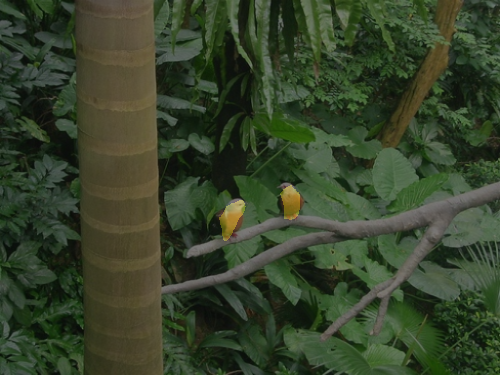}
    \includegraphics[width=0.24\textwidth]{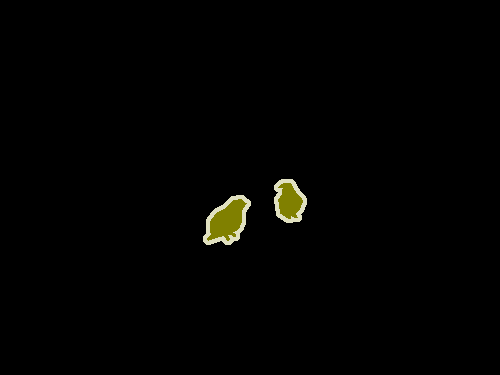}
    \\ 
    \begin{tabular}{ >{\centering}m{0.22\textwidth} >{\centering}m{0.22\textwidth} >{\centering}m{0.22\textwidth} >{\centering}m{0.22\textwidth}}
     Image & MobileViT-S-DeepLabV3~\cite{mobilevit_iclr2022} & MobileOne-S4-DeepLabV3 (ours) & Ground Truth \\
    \end{tabular}
    \includegraphics[width=0.8\textwidth]{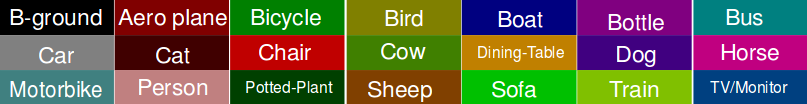}
    \caption{Qualitative results on semantic segmentation. Legend reproduced from DeepLab~\cite{deeplabv3}.}
        \label{fig:semantic_segmentation}        
\end{figure*}

\section{Semantic Segmentation}
\subsection{Training details}
We use the MobileViT repository~\cite{mobilevit_iclr2022} to train our semantic segmentation models and adopt their hyperparameter settings.
Both VOC and ADE20k segmentation models were trained for 50 epochs using cosine learning rate with a maximum learning rate of $10^{-4}$ and minimum learning rate of $10^{-6}$. We use 500 warmup iterations. The segmentation head has a learning rate multiplier of 10. EMA is used with a momentum of $5 \times 10^{-4}$. We use AdamW optimizer \cite{adamw} with weight decay of 0.01. For VOC, the model is trained on both MS-COCO and VOC data simultaneously following Mehta et al~\cite{mobilevit_iclr2022}. For both VOC and ADE20k, the only augmentations used are random resize, random crop, and horizontal flipping. 

\subsection{Qualitative Results}
We provide qualitative results for semantic segmentation in Figure~\ref{fig:semantic_segmentation}. Our method performs better than MobileViT-S-DeepLabV3 as shown. In row 1, we show that MobileViT-S misclassifies background as airplane. In row 2 and row 6, our method is able to resolve fine details such as the leg of the horse and tiny birds. In row 3, MobileViT-S misclassfies the couch. In row 4, our method is able to segment large foreground object at a close-up view. In row 5, our method segments small objects such as the buses.

\end{document}